\documentclass[conference]{IEEEtran}
\usepackage[utf8]{inputenc} %
\usepackage[T1]{fontenc}    %
\PassOptionsToPackage{hyphens}{url}
\usepackage[colorlinks]{hyperref}
\usepackage{booktabs}       %
\usepackage{amsfonts}       %
\usepackage{nicefrac}       %
\usepackage{microtype}      %
\usepackage{color}
\usepackage{refstyle}
\usepackage{amsmath}
\usepackage{amssymb}
\usepackage{amsthm}
\usepackage{bm}
\usepackage{graphicx}
\usepackage{textcomp}
\usepackage{xcolor}
\usepackage{color, colortbl}
\usepackage{float}
\usepackage{threeparttable, tablefootnote}
\usepackage{epstopdf}
\usepackage{multicol}
\usepackage{multirow}
\usepackage{subfigure}
\usepackage{bbm}
\usepackage[bottom,hang]{footmisc}
\usepackage{adjustbox}
\usepackage{wrapfig}
\usepackage{caption}
\usepackage[normalem]{ulem}
\usepackage{arydshln}
\usepackage[linesnumbered,lined,vlined,ruled]{algorithm2e}
\usepackage{xpatch}
\usepackage{dsfont}
\usepackage{tabularx}
\usepackage{comment}
\let\labelindent\relax
\usepackage{enumitem}
\usepackage{mathtools}
\def\eqref#1{equation~\ref{#1}}
\def\1{\bm{1}}

\def\vtheta{{\bm{\theta}}}

\def\vs{{\bm{s}}}

\def\vx{{\bm{x}}}

\def\mI{{\bm{I}}}

\DeclareMathAlphabet{\mathsfit}{\encodingdefault}{\sfdefault}{m}{sl}
\SetMathAlphabet{\mathsfit}{bold}{\encodingdefault}{\sfdefault}{bx}{n}

\def\gA{{\mathcal{A}}}

\def\gM{{\mathcal{M}}}
\def\gN{{\mathcal{N}}}

\def\gS{{\mathcal{S}}}
\def\gT{{\mathcal{T}}}

\def\sS{{\mathbb{S}}}

\newcommand{\E}{\mathbb{E}}
\newcommand{\Ls}{\mathcal{L}}

\newcommand{\KL}{D_{\mathrm{KL}}}

\DeclareMathOperator*{\argmin}{arg\,min}

\theoremstyle{definition}

\xpretocmd{\proof}{\setlength{\parindent}{0pt}}{}{}

\newcommand{\myparatight}[1]{\smallskip\noindent{\bf {#1}:}~}

\newcommand{\myparagraph}[1]{
\vspace{0.1cm}\noindent
\textbf{#1.}
}

\renewcommand\sectionautorefname{Section}

\setitemize{itemsep=0pt,leftmargin=*}
\setenumerate{itemsep=0pt,leftmargin=*}

\usepackage{tikz}
\usepackage{amsmath}

\usepackage{filecontents}

\usepackage{graphicx}

\newcommand{\dingfan}[1]{\textcolor{blue}{{\it [Dingfan: #1]}}}

\makeatletter
\newcommand{\linebreakand}{%
  \end{@IEEEauthorhalign}
  \hfill\mbox{}\par
  \mbox{}\hfill\begin{@IEEEauthorhalign}
}
\makeatother

\ifCLASSINFOpdf
\else
\fi
\usepackage{comment}

\begin{document}
\title{\Large \bf Data Forensics in Diffusion Models: A Systematic Analysis of Membership Privacy }

\author{\IEEEauthorblockN{Derui Zhu\textsuperscript{$\ast$}}
\IEEEauthorblockA{Technical University of Munich\\
derui.zhu@tum.de}
\and
\IEEEauthorblockN{Dingfan Chen\textsuperscript{$\ast$}}
\IEEEauthorblockA{CISPA Helmholtz Center for Information Security\\
dingfan.chen@cispa.de}
\linebreakand
\IEEEauthorblockN{Jens Grossklags}
\IEEEauthorblockA{Technical University of Munich\\
jens.grossklags@in.tum.de}
\and 
\IEEEauthorblockN{Mario Fritz}
\IEEEauthorblockA{CISPA Helmholtz Center for Information Security\\
fritz@cispa.de}}


%\author{\IEEEauthorblockN{Anonymous Authors}
%\IEEEauthorblockA{\\}
%}


\IEEEoverridecommandlockouts
\makeatletter\def\@IEEEpubidpullup{6.5\baselineskip}\makeatother
\IEEEpubid{\parbox{\columnwidth}{
     $\ast$: equal contributions
}
\hspace{\columnsep}\makebox[\columnwidth]{}}% make the title area
\maketitle

\begin{abstract}
%\boldmath
In recent years, diffusion models have achieved tremendous success in the field of image generation, becoming the state-of-the-art technology for AI-based image processing systems. Despite the numerous benefits brought by recent advances in diffusion models, there are also concerns about their potential misuse, specifically in terms of privacy breaches and intellectual property infringement.
In particular, some of their unique characteristics open up new attack surfaces when considering the real-world deployment of systems built on such models. With a thorough investigation of the attack vectors, we develop a systematic analysis of membership inference attacks on diffusion models and propose novel attack methods tailored to each attack scenario specifically relevant to diffusion models.
Our approach exploits easily obtainable quantities and is highly effective, achieving near-perfect attack performance ($>$0.9 AUCROC) in realistic scenarios. Our extensive experiments demonstrate the effectiveness of our method, highlighting the importance of considering privacy and intellectual property risks when using diffusion models in image generation systems. 
\end{abstract}

\section{INTRODUCTION}
\label{sec:intro}
%-------------------------------------------------------------------------------
Deep generative modeling has made significant advancements over the past few years, resulting in photo-realistic media generation tools with emerging commercial uses for art and design. In particular, the rapid improvement of denoising diffusion models~\cite{sohl2015deep,nichol2021improved,ho2020denoising,song2019generative,song2020improved,songscore,dhariwal2021diffusion} has greatly advanced the state-of-the-art in the image and video generation tasks, as highlighted in recent studies~\cite{dhariwal2021diffusion}. Meanwhile, diffusion models are considered the most promising generative framework to date, serving as the foundation for powerful commercial models such as Stable Diffusion~\cite{rombach2022high}, Imagen~\cite{saharia2022photorealistic}, and DALL$\cdot$E-2~\cite{ramesh2022hierarchical}.

Despite the remarkable success of recent diffusion models, the widespread use of online APIs and shared pre-trained models raises concerns about their potential risks in various areas. One major concern is the risk of data misuse and violations of privacy, as sensitive information pertaining to individual identities could be revealed. Additionally, malicious users may attempt to infer the original training data, further exacerbating privacy concerns. An example of such an attack is the membership inference attack (MIA)~\cite{shokri2017membership}, which seeks to determine if a particular data record was used to train a machine learning model. This is particularly concerning in the context of diffusion models that serve as the backbone for online media editing tools, which are freely accessible to the public.

Another major concern is the potential intellectual property (IP) infringement during the development and deployment of diffusion models. Advanced diffusion models rely heavily on the usage of massive and diverse training data. However, with the commercialization of these models, there is a risk of data being harvested from the internet for model training purposes without proper regard for the IP rights of media creators. Most commonly, it is impractical for model developers to manually review all training samples for IP compliance, making this a relevant issue in real-world application scenarios.

Diffusion models possess several distinct features that set them apart from other generative models. First of all, the encoding process in diffusion models is unlearnable and fixed, following a standard procedure that is known to the public. While this eases the training of diffusion models on complex data distributions, it also presents vulnerabilities as attackers can easily and precisely imitate the encoding process, even if the model developer tries to hide it during model deployment (\sectionautorefname~\ref{subsec:adpative_defense}). In contrast, attacks on other generative models usually require approximating the encoding process in a lossy manner, e.g., through gradient-based optimization on the model internals~\cite{chen2020gan} (\sectionautorefname~\ref{subsec:diss_comparison_gm}). Additionally, the generation process in diffusion models is iterative, resulting in multiple intermediate outputs that may all reveal information about the training samples. Such information can be easily exploited by an attacker to construct dedicated attacks tailored to diffusion models under different deployment scenarios.

% Note that diffusion models have the unique advantage of generating images progressively, differentiating them from GAN models. Previous studies~\cite{chen2020gan,hilprecht2019monte} have shown that their approaches heavily rely on  the similarity between generated images and the original training images. However, in diffusion models, adversaries are able of exploiting additional information to MIA, e.g. images generated in different steps.  \mario{we are missing a paragraph here that makes it clear - why prior work is insufficient to address membership inference for diffusion models. right now the differences between membership inference on GAN vs difussion are pushed down deep into section 4. this is a strategic mistake I think. we should do the opposite and pull it up into intro - even abstract! what is the key insight? why can't we apply MI from GAN directly on diffusion models? what insight are we exploiting.}

In this work, we pioneer the investigation of such risks associated with diffusion models.  Specifically, we conduct the first systematic analysis of MIAs against diffusion models. 
While previous studies have explored MIAs in the context of both classification models~\cite{shokri2017membership,ndss19salem,carlini2022membership,sablayrolles2019white,yeom2018privacy,song2021systematic,nasr2019comprehensive} and other generative models~\cite{hayes2019logan,hilprecht2019monte,chen2020gan}, we highlight that diffusion models have unique properties and usage patterns that create new attack surface not covered by existing works. Furthermore, existing methods are not directly applicable to our scenario, while even potential adaptations would only yield suboptimal special cases of our proposed attack (refer to \sectionautorefname~\ref{subsec:analytical_insights}).  Instead, we thoroughly examine the attack vectors and identify three attack scenarios that are most representative and prevalent in practice, given real-world APIs as reference. Moreover, we design novel attacks tailored to diffusion models based on their unique characteristics, achieving near-perfect performance ($>$$0.9$ AUCROC) across various practical settings.

% Lastly, we relate our approach to the protection of intellectual property rights during the deployment of diffusion models. Model developers can provide assurance to requesters that a specific query sample was not used during training, as indicated by high confidence in predicting the query to be non-members by our attacks. In the event of suspected misuse, our MIAs confidence score can be used for additional verification.
%-------------------------------------------------------------------------------
%\subsubsection*{Contributions}
\noindent\textbf{Contributions:}
In summary, we make contributions on three distinct levels in this paper, which we categorize as task-level, approach-level, and insight-level.
\begin{itemize}
    \item \myparatight{Task-level} We carry out the first systematic investigation of membership inference attacks on state-of-the-art diffusion models. With a  thorough analysis of the potential attack surface, our study
reveals the most realistic threat models that reflect actual usage patterns and categorizes attack scenarios into white-box, gray-box, and black-box settings, depending on the information available to the attacker. 
These categorizations have high practical relevance, reflecting common real-world scenarios while guiding and benchmarking future research in related fields.

\item \myparatight{Approach-level}
 We design novel attack strategies for diffusion models, customized to suit various scenarios. Our attacks are based on easily obtainable or estimable quantities and are both straightforward and highly effective, supported by a theoretical basis. 
 Moreover, our proposed improvement techniques (e.g., truncation and calibration as discussed in Sections~\ref{subsec:whitebox} and \ref{subsec:greybox}) significantly enhance attack performance in realistic settings and are highly practical. We anticipate that our proposed method and insights will have broader applications beyond membership inference and be of interest for future tasks involving diffusion models.  

 \item \myparatight{Insight-level}
  We conduct a thorough evaluation of our proposed attack, taking into account various factors such as attack scenarios, data distributions, sample size, target model, and training configurations. We provide a detailed analysis of the key components that could impact the effectiveness of our attack. We find that our approach is consistently effective across different scenarios. Specifically, having only access to the API, our approach reaches $>$$0.95$ AUCROC on the CelebA dataset with 20k training samples, where previous work generally fails to report effective attacks. Moreover, our attack demonstrates substantial effectiveness, evidenced by an AUCROC of $>$$0.7$ and a $>$$24\%$ TPR@$1\%$ FPR, when applied to real-world pre-trained models like Stable-Diffusion trained on large-scale datasets containing  $2.3$ billion  samples. 
  Our findings point to a two-fold implication: on the one hand, there is an exceptionally high privacy risk associated with the common practice of sharing diffusion models; on the other hand, our attack strategy holds the potential to serve as foundational elements for cases that necessitate the monitoring of sample usage during the training of a diffusion model,  serving purposes like IP protection.   %Lastly, our results suggest, on the one hand, an exceptionally high privacy risk of sharing diffusion models as a common practice, while on the other hand, the potential of our attack to be used for faithfully tracking the usage of certain samples during training of a diffusion model for the sake of e.g., IP protection. %1.0 AUCROC in CelebA dataset with 10k training samples. When the training sample size increases to 20k, our approach remains effective (attack AUC > 0.95). 
\end{itemize}

\section{RELATED WORK}
\label{sec:related_work}
%-------------------------------------------------------------------------------
\myparagraph{Generative Models}
 Generative models aim to simulate the probability distribution of real data by defining a parametric family of densities and finding the optimal parameters. The optimal parameter is typically found by either maximizing the (lower bound of) likelihood of the real data or minimizing the (estimated) divergence between the generated and real data distributions. With the advancement in the expressive power of deep neural networks, recent generative models have achieved significant success in modeling high-dimensional data distributions. Different types of deep generative models have been developed in the literature, with generative adversarial networks (GANs)~\cite{NIPS2014_5ca3e9b1}, variational autoencoders (VAEs)~\cite{kingma2014autoencoding}, and diffusion-based models~\cite{sohl2015deep,song2019generative,ho2020denoising} being the representative ones. \textit{GANs} are made up of two modules: the generator and the discriminator, which are trained jointly in a competitive manner. The discriminator is trained to estimate the divergence between the generated and real data distributions, while the generator is optimized to reduce this divergence so as to closely resemble the real data distribution. 
 In contrast, \textit{diffusion models} and \textit{VAEs} are trained based on a log-likelihood objective, , adhering to the variational autoencoding pipeline~\cite{kingma2014autoencoding}. This involves mapping images to a (Gaussian) latent space, then transforming them back into the data space with a learnable decoder. While diffusion models employ a predefined, unlearnable forward encoding process coupled with an iterative decoding process, VAEs learn the encoding process and implement a one-time, non-iterative decoding process.

 In this work, we focus on diffusion models, which represent the current state-of-the-art deep generative framework~\cite{dhariwal2021diffusion} and serve as the backbone for various online media generation tools~\cite{saharia2022photorealistic,rombach2022high,ramesh2022hierarchical}. Additionally, we make connections and draw comparisons with GANs and VAEs (\sectionautorefname~\ref{subsec:exp_different_gm} and
\ref{subsec:diss_comparison_gm}), which were the previous leading generative frameworks. 

\myparagraph{Membership Inference Attacks (MIAs)}
Membership Inference Attack (MIA) was first introduced by Shokri et al.~\cite{shokri2017membership}. It focuses on attacking classification models in a black-box setting, where the attacker has access to the victim model's full response, including confidence scores for all classes, for a given query 
 sample as input. Existing works have developed various approaches in attacking both white-box~\cite{nasr2019comprehensive,rezaei2020towards} as well as black-box~\cite{shokri2017membership,sablayrolles2019white,yeom2018privacy,ndss19salem,song2021systematic} classification models. 
 Black-box MIAs typically involve training shadow models to extract member and non-member characteristics or utilizing easily accessible information such as losses as the membership indicator. White-box attacks, on the other hand, utilize the target model's internals (e.g., sample gradients) to construct membership scores.
 In particular, 
 it has been shown that the sample loss generally can serve as a discriminative signal that tells apart members from non-members~\cite {yeom2018privacy}.
 Sablayrolles et al.~\cite{sablayrolles2019white} further showed that black-box attacks can approximate the performance of white-box MIA under certain assumptions on the model parameter distribution. Our approach is built on top of these findings and is specifically tailored for diffusion models by carefully examining their training objectives and potential attack surface during development and deployment.

Recent works have explore such attacks for popular generative models such as GANs~\cite{hayes2019logan,chen2020gan} and VAEs~\cite{hilprecht2019monte}. Specifically, Hayes~et~al.~\cite{hayes2019logan} observe that disclosing the discriminator in a GAN can result in leaked membership information in a white-box setting and suggested using a shadow model for black-box attacks.
Hilprecht~et~al.~\cite{hilprecht2019monte} proposed using the reconstruction error as a membership score for attacking white-box VAEs and counting generated samples within an $\epsilon$-ball of the query for a black-box membership score. Chen et al.~\cite{chen2020gan} presented a taxonomy of MIAs against GANs and proposed an optimization-based approach for attacks with only generator access and a distance-based approach for the black-box setting with only synthetic samples available.

Our work presents the first systematic analysis of MIAs on diffusion models. Despite similarities in the training objectives with VAEs and comparable generation quality to GANs, diffusion models have distinct properties and unique attack vectors  that can be considered and exploited by attackers. We thoroughly examine different attack scenarios specifically relevant to diffusion models and leverage its intrinsic characteristics, such as the pre-defined encoding process and multi-step generation process, to conduct effective attacks. Algorithmically, our approach shares the same high-level concept with existing sample loss-based techniques~\cite{yeom2018privacy,sablayrolles2019white,chen2020gan}, but differs fundamentally by exploiting the intrinsic properties of diffusion models to make the membership score representative and discriminative, leading to exceptional performance.
Notably, while there have been some very recent attempts, mostly unpublished, to investigate privacy attacks on diffusion models~\cite{kong2023efficient,wu2022membership,matsumoto2023membership,hu2023membership,carlini2023extracting,duan2023diffusion}, these efforts only constitute a subset of the scenarios investigated in our work, with their proposed attacks generally fall into special (and sub-optimal) cases of our attack model. Specifically, \cite{wu2022membership,kong2023efficient} corresponds to the black-box setting in our study, whereas \cite{duan2023diffusion,hu2023membership} aligns with our gray-box setting, and \cite{carlini2023extracting,matsumoto2023membership} focus on the white-box and black-box settings. 

% \myparagraph{Threats Related to Generative Models}

%-------------------------------------------------------------------------------
\section{BACKGROUND}
\label{sec:background}

\subsection{Diffusion Models}
\myparagraph{Formulation} Given observed samples $\vx_0$ from a distribution of interest, the goal of a generative model is to learn to model its true underlying distribution $p(\vx_0)$ and generate novel samples from it. Specifically, diffusion models use a forward noising process $q$, i.e., the ``encoding'' process, to gradually transform the data distribution into a standard Gaussian $\gN(0,\mI)$.  The models then learn to reverse this transformation through a learnable denoising function $p_\vtheta$, i.e., the "decoding" process. Once the denoising function $p_\vtheta$ is learned, generating new samples from the data distribution can be achieved by sampling from the standard Gaussian and then iteratively applying the reverse denoising steps $p_\vtheta(\vx_{t-1}|\vx_t)$. Formally, the forward noising process can be written as follows,
 \begin{equation}
 \label{eq:forward}
    q(\vx_t | \vx_{t-1} ) = \gN (\vx_t ;\sqrt{\alpha_t} \vx_{t-1} , (1 -\alpha_t)\mI)
\end{equation}
where the subscript $t$ is the step index, and  $\alpha_t$ is a scaling factor ($0\leq \alpha_t \leq 1$) controlling the amount of information preserved in each noising step (where a larger $\alpha_t$ means more information is kept). Given a sufﬁciently large $T$ and an appropriate schedule of $\alpha_T$, the latent $\vx_T$ at the final step forms a standard Gaussian distribution. Meanwhile, the forward process defined in \equationautorefname~\ref{eq:forward} allows direct sampling of the noisy latent $\vx_t$ at an arbitrary step given the input data $\vx_0$~\cite{ho2020denoising}:
\begin{align}
q(\vx_t|\vx_0) &= \gN(\vx_t;\sqrt{\bar{\alpha}_t}\vx_0, (1-\bar{\alpha}_t)\mI) \label{eq:q_xt}\\
\vx_t &=\sqrt{\bar{\alpha}_t}\vx_0 + \sqrt{1-\bar{\alpha}_t}\epsilon
\label{eq:x_t}
\end{align}
where $\bar{\alpha}_t = \prod_{s=0}^t \alpha_s$ and  $\epsilon \sim \gN(0,\mI)$ denotes a random noise sample.
Moreover, the posterior $q(\vx_{t-1}|\vx_t,\vx_0)$ can be computed using Bayes theorem:
\begin{align}
    q(\vx_{t-1}|\vx_t,\vx_0) &= \gN(\vx_{t-1};\mu_q (\vx_t,\vx_0),\Sigma_q(t)) \label{eq:posterior}\\
    \mu_q(\vx_t,\vx_0)&= \frac{\sqrt{\alpha_t}(1-\bar{\alpha}_{t-1})\vx_t+\sqrt{\bar{\alpha}_{t-1}}(1-\alpha_t)\vx_0}{1-\bar{\alpha}_t} \nonumber \\
    \Sigma_{q}(t) &=  \frac{(1-\alpha_t)(1-\bar{\alpha}_{t-1})}{1-\bar{\alpha}_t}\mI \nonumber
\end{align}

\noindent The joint distribution for the reverse process can be formularized as:
\begin{equation}
    p(\vx_{0:T} ) = p(\vx_T) \prod_{t=1}^T p_\vtheta(\vx_{t-1}|\vx_t ) 
\end{equation}
with $p(\vx_T) = \gN(\vx_T;0,\mI)$, indicating that the latent distribution at the final step $T$ is a standard Gaussian. The denoising function is modeled as a Gaussian using a neural network as follows:
\begin{equation}
p_\vtheta (\vx_{t-1}|\vx_t) = \gN(x_{t-1}; \mu_\vtheta(\vx_t, t),\Sigma_\vtheta(\vx_t,t) )
\label{eq:p_theta}
\end{equation}
% with the neural network predicting the mean $\mu$ and covariance $\Sigma$ of a Gaussian distribution given the noisy latent $\vx_t$ and time step $t$ as input. 

\myparagraph{Objective}
The diffusion models are trained to maximize the variational lower bound (VLB), i.e., a lower bound of the log-likelihood of the observed data. Formally, 
\begin{align*}
     \log p(\vx_0) %&= \log \int p(\vx_{0:T}) d\vx_{1:T} \\
    % &= \log \int \frac{p(\vx_{0:T})q(\vx_{1:T}|\vx_0)}{q(\vx_{1:T}|\vx_0)} d\vx_{1:T} \\
    & \geq \E_{q(\vx_1|\vx_0)}[\log p_\vtheta(\vx_0|\vx_1)] - \KL(q(\vx_T|\vx_0)\Vert p(\vx_T)) \\
    &- \sum_{t=2}^T\E_{q(\vx_t|\vx_0)}[\KL(q(\vx_{t-1}|\vx_t,\vx_0)\Vert p_{\vtheta}(\vx_{t-1}|\vx_t))]
\end{align*}
where $\KL$ denotes the KL divergence.
The training objective can be equivalently written as minimizing the negative VLB:
\begin{align}
        \vtheta^* &= \argmin_{\vtheta} \Ls_{vlb}= \argmin_{\vtheta} \{\Ls_0 + \Ls_1 +... +\Ls_T\} \label{eq:objective_total} \\
        \Ls_0 &= -\log p_\vtheta(\vx_0|\vx_1) \\
        \Ls_{t-1} &= \KL(q(\vx_{t-1}|\vx_t,\vx_0)\Vert p_{\vtheta}(\vx_{t-1}|\vx_t)) ,\, 2\leq t\leq T \label{eq:objective_t}\\
        \Ls_T &= \KL (q(\vx_T|\vx_0)\Vert p(\vx_T))
\end{align}
In practice, $\Ls_0$ is computed by discretizing each color component into 256 bins, and evaluating the probability of $p_\vtheta (\vx_0 | \vx_1 )$ landing in the correct bin~\cite{dhariwal2021diffusion,nichol2021improved}.  $\Ls_{t-1}$ in \equationautorefname~\ref{eq:objective_t} is computed by sampling from an arbitrary step of the forward noising process (\equationautorefname~\ref{eq:x_t}) and estimate $\Ls_{t-1}$ using \equationautorefname~\ref{eq:posterior} and \ref{eq:p_theta}. Optimizing over the sum in \equationautorefname~\ref{eq:objective_total} on the training dataset is achieved by randomly sampling $t$ for each image  $\vx_0$ in each mini-batch, i.e.,  approximating $\Ls_{vlb}$ using the expectation $\E_{t,\vx_0,\epsilon}[\Ls_{t}]$.

\myparagraph{Parameterization}
Recent works develop different ways to parameterize $p_\vtheta$ in \equationautorefname~\ref{eq:p_theta} for solving $\argmin_\vtheta \Ls_{t-1}$:
\begin{align}&\argmin_\vtheta \Ls_{t-1} \nonumber \MoveEqLeft[1]\\
&= 
    \argmin_\vtheta  \frac{1-\bar{\alpha}_t}{2(1-\alpha_t)(1-\bar{\alpha}_{t-1})} \Vert \mu_\vtheta(\vx_t,t) -\mu_q \Vert_2^2 \label{eq:predict_mean}\\ %\KL(q(\vx_{t-1}|\vx_t,\vx_0)\Vert p_\vtheta(\vx_{t-1}|\vx_t)) \\
    &= \argmin_\vtheta  \frac{\bar{\alpha}_{t-1}(1-\alpha_t)}{2(1-\bar{\alpha}_t)(1-\bar{\alpha}_{t-1})}\Vert \hat{\vx}_\vtheta(\vx_t,t)-\vx_0\Vert_2^2 \label{eq:predict_x}\\
    &= \argmin_\vtheta \frac{(1-\bar{\alpha}_t)(1-\alpha_t)}{2\alpha_t(1-\bar{\alpha}_{t-1})}\Vert \epsilon_0 - \hat{\epsilon}_\vtheta (\vx_t,t)\Vert_2^2 \label{eq:predict_noise}\\
    &= \argmin_\vtheta \frac{(1-\bar{\alpha}_t)(1-\alpha_t)}{2\alpha_t(1-\bar{\alpha}_{t-1})} \resizebox{0.42\columnwidth}{!}{$ \Vert s_\vtheta(\vx_t,t)-\nabla \log p(\vx_t)\Vert _2^2
    $}\label{eq:predict_score}
\end{align}
The most obvious option is to let the neural network predict $\mu_\vtheta(\vx_t,t)$ directly (\equationautorefname~\ref{eq:predict_mean}). Alternatively, the network could predict $\vx_0$ from noisy image $\vx_t$ and time index $t$ (\equationautorefname~\ref{eq:predict_x})~\cite{ho2020denoising}. The network could also predict the noise $\epsilon_0$ that determines $\vx_t$ from $\vx_0$ (\equationautorefname~\ref{eq:predict_noise})~\cite{saharia2022photorealistic,ho2020denoising} and the score of the image at an arbitrary noise level, i.e, the gradient of $\vx_t$ in data space (\equationautorefname~\ref{eq:predict_score})~\cite{song2019generative,song2020improved,songscore}.

Our attack is designed based on the general objective and quantities that all diffusion models rely on, making it easily applicable to existing models and seamlessly adaptable to future advancements (See \sectionautorefname~\ref{sec:general_attack_pipeline} for more details.)

\subsection{Membership Inference}
\label{sec:background_MI}
In the standard setting of MIA, the attacker has access to a query set  $\sS=\{(\vx^i, m^i)\}_{i=1}^N$ with both member (training) and non-member (testing) samples $\vx^i$ drawn from the same distribution,  where $m^i$ is the membership attribute ($m^i=1$ if $\vx^i$ is a member). Unless stated otherwise, the sample index is denoted by the superscript $i$, and the time step index in a diffusion model is denoted by the subscript $t$. 
The task is to determine the membership attribute $m^i$ of each sample $\vx^i$. By default, we treat each image as a separate sample. For conditional generation tasks, a sample may encompass additional features, such as text descriptions. In such case, we write $\vx^i=(\vx^i_{\mathrm{img}},\vx^i_{\mathrm{text}})$ to avoid confusion.
The attack $\gA(\vx^i,\gM(\vtheta))$ is a binary classifier that predicts $m^i$ for a given  query sample $\vx^i$ and a target model $\gM$ parametrized by $\vtheta$, i.e., a diffusion model with model parameter $\vtheta$. The Bayes optimal attack $\gA_{opt}(\vx^i,\gM(\vtheta))$ will output 1 if the query sample is more likely to be contained in the training set, based on the real underlying membership probability %
$P(m^i=1|  \vx^i, \vtheta)$, which is usually formulated as a non-negative log ratio:
\begin{equation}
\label{eq:opt_ratio}
    \mathcal{A}_{opt}(\vx^i,\gM(\vtheta)) = \mathds{1}  \left[ \log \frac{P(m^i=1|  \vx^i, \vtheta)}{P(m^i=0|  \vx^i, \vtheta)} \geq 0 \right]
\end{equation}
with $\mathds{1} $ denoting the indicator function.

Our attack is motivated by the recent results showing the dependence of attack success rate on the sample loss~\cite{yeom2018privacy,sablayrolles2019white}. In particular, a large difference in expected loss values between the member and non-member data has been shown to be a sufficient condition for conducting successful attacks~\cite{yeom2018privacy}. 
Sablayrolles et al.~\cite{sablayrolles2019white} further prove that the Bayes optimal attack depends only on the sample loss under a mild posterior assumption of the model parameter, expressed as $\mathcal{P}(\vtheta|\vx^1,..,\vx^n)\propto e^{ -\frac{1}{\gT} \sum_{i=1}^N m^i \cdot \ell(\vtheta,\vx^i)}$. This corresponds to a Bayesian perspective in which $\vtheta$ is viewed as a random variable that minimizes the empirical risk $\sum_{i=1}^N m^i \cdot \ell(\vtheta,\vx^i)$, and $\gT$ is the temperature parameter representing the stochasticity in the model. The optimal attack is then formularized as follows: 
\begin{equation}
\label{eq:opt_attack}
    \mathcal{A}_{opt} (\vx^i,\gM(\vtheta)) = 
\mathds{1}\big[\ell(\vtheta,\vx^i)<\tau(\vx^i)\big]
\end{equation} where $\tau$ denotes a threshold function. 

Intuitively, \equationautorefname~\ref{eq:opt_attack} indicates that $\vx^i$ is more likely to be a training sample if the target model shows a low loss value for it. While many existing attacks~\cite{chen2020gan,sablayrolles2019white,hilprecht2019monte} fall into this framework, simply applying these results to attacking diffusion models would suggest using the VLB loss $\Ls_{vlb}$ (\equationautorefname~\ref{eq:objective_total}) as the membership score (see more discussions in \sectionautorefname~\ref{subsec:analytical_insights}). However, as demonstrated in Sections~\ref{sec:general_attack_pipeline} and \ref{sec:experiments}, this approach may result in sub-optimal performance or even be infeasible under realistic threat models. Our proposed approach, tailored for each possible threat model, is described in detail in \sectionautorefname~\ref{sec:general_attack_pipeline}.
\section{GENERAL ATTACK PIPELINE}
\label{sec:general_attack_pipeline}
Our attack draws inspiration from previous findings that the sample loss ($\Ls_{vlb}$ in our case) can serve as a reliable indicator for membership (\sectionautorefname~\ref{sec:background_MI}). However, we uncover several limitations with a straightforward approach and propose modifications tailored to specific threat models. We summarize our threat model in \tableautorefname~\ref{table:taxonomy} and present our proposed improved approach in the upcoming Sections~\ref{subsec:whitebox}-\ref{subsec:blackbox}. 

 \begin{figure*}[!t]
\centering
\subfigure[Distribution of $\Ls_{t}$]{
\includegraphics[width=0.48\linewidth]{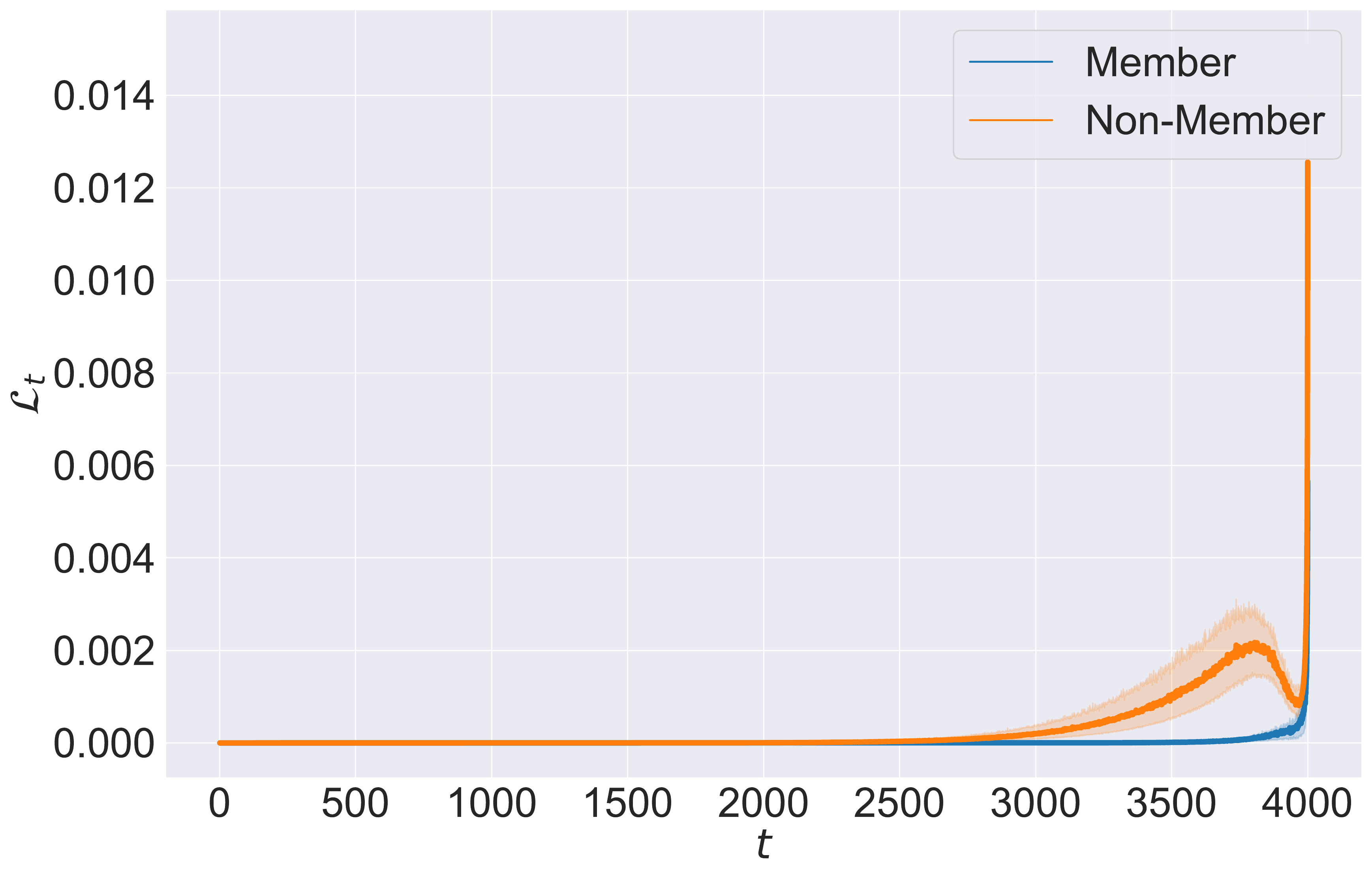}
\label{fig:whitebox_vb_trajactory}
}
\hfill
\subfigure[Distribution of $\widehat{\Ls_{t}}$]{\includegraphics[width=0.48\linewidth]{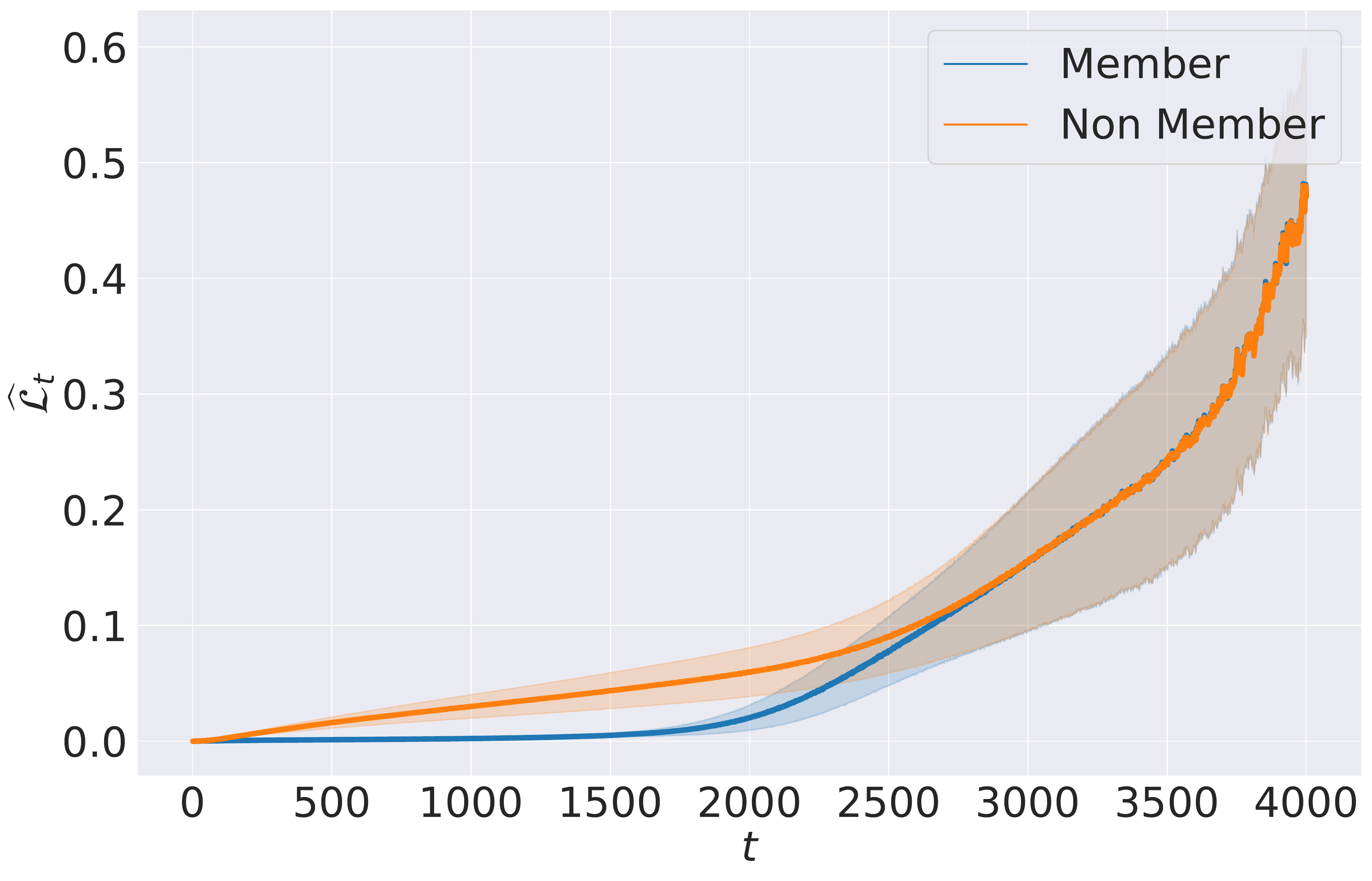}
\label{fig:whitebox_xstart_mse_trajactory}
}
\centering
\caption{The distribution of $\Ls_{t}$ (\ref{fig:whitebox_vb_trajactory}) and $\widehat{\Ls_{t}}$ (\ref{fig:whitebox_xstart_mse_trajactory}) terms for each reverse denoising step $t$ of a target diffusion model trained on CelebA with 5k samples.}
\label{fig:trajactory}
\end{figure*}
\subsection{White-box setting}
\label{subsec:whitebox}
\myparagraph{Threat Model}
We start by investigating the most informed attacker scenario, which is the white-box setting. In this scenario, the attacker has complete access to the trained model parameters $\vtheta$, as well as the necessary information for model implementation, such as the number of total steps $T$ as well as the (schedule of) scaling factor $\alpha_t$ used in the forward and backward pass. This scenario reflects the common open-source practice of releasing the source code and the pre-trained model checkpoints for public use. These resources, often used as building blocks for image editing tools, are readily available on the internet. 

\myparagraph{Approach}
Existing results suggest that utilizing the sample loss is a viable approach for calculating membership score (see \sectionautorefname~\ref{sec:background_MI}). However, relying solely on $\Ls_{vlb}$ results in subpar outcomes (see \sectionautorefname~\ref{subsec:eval_white_box}). We conjecture that this is mainly due to the following reasons: First, the randomness in the sampling process during training may cause unequal weight of each term $\Ls_t$ in the total sum, leading to a deviation from the intended objective $\Ls_{vlb}$. More importantly, the large variance in the scale of each term $\Ls_t$ results in less informative terms dominating the $\Ls_{vlb}$ sum, leading to a less discriminative outcome for membership inference. As depicted in \figureautorefname~\ref{fig:whitebox_vb_trajactory}, the terms closer to the noise end (i.e., with a large value of $t$) have the greatest impact on the VLB loss, but they may be less informative in determining membership. This is because they are close to the Gaussian noise endpoint $\vx_T$ and contain limited information about the sample itself $\vx_0$. 

Hence, we propose using the independent terms $\Ls_t$ that are most discriminative for membership inference instead of $\Ls_{vlb}$ (the sum of all terms). For this, we focus on terms $0 \leq t \leq T_\mathrm{trun}$, where $T_\mathrm{trun}$ is the point where the loss terms become significantly larger than the previous ones. A simple rule of thumb is to set $T_\mathrm{trun}$ to approximately $0.75T$, which achieves high effectiveness and is not sensitive to different datasets. In practice, a reasonable choice of $T_\mathrm{trun}$ can be selected on a small reference set. We also investigate various statistical functions to summarize the loss terms (expressed as a general function $f$ in \equationautorefname~\ref{eq:white_box_attack}) instead of just the sum as in $\Ls_{vlb}$ (see \sectionautorefname~\ref{subsec:eval_white_box}). We anticipate that a learnable function $f$ may also be effective, but for simplicity and high attack effectiveness, we use simple statistics such as mean and median. Our attack can be formulated as follows: 
\begin{equation}
\label{eq:white_box_attack}
\mathcal{A}(\vx^i,\gM(\vtheta)) = 
\mathds{1}\Big[f\Big(\big\{\Ls_{t}(\vtheta,\vx^i)\big\}_{t=0}^{T_\mathrm{trun}}\Big)<\tau(\vx^i)\Big]
\end{equation}
In line with existing results, the attack predicts that a sample belongs to the training set if its overall loss (summarized by $f$) is lower than a threshold $\tau(\vx^i)$. The threshold function can be calibrated to each sample to account for the impact of sample difficulty on membership inference~\cite{sablayrolles2019white,carlini2022membership}.

% \begin{figure}[!t]
% \centering
% \includegraphics[width=\linewidth]{figures/whitebox_xstart_mse_trajectory_celeba.pdf}
% \caption{ }
% \label{fig:whitebox_xstart_mse_trajactory}
% \end{figure}

\subsection{Gray-box setting}
\label{subsec:greybox}
\myparagraph{Threat Model}
In this setting, we consider a more realistic scenario where the attacker does not have direct access to the model parameter $\vtheta$, but can still execute the model. This closely resembles a situation where the model is available to the public through an online API, where the model owner allows others to use the essential functions of the model without disclosing the underlying model. The information that the attacker can exploit may vary depending on the information released through the API. For example, some APIs allow greater control over the generation process, while others do not. Here, we present our attack designed for scenarios that most closely resemble existing real-world APIs and defer the discussion of more relaxed settings to the next subsection.

\myparagraph{Approach} Similar to attacking a white-box diffusion model, we still use the sample loss as the membership indicator. However, the attacker does not have direct access to the terms $\Ls_{t}$. Instead, what the attacker can access are the intermediate outputs $\hat{\vx}_\vtheta(\vx_t,t)$ of the diffusion models applying $t$ denoising steps given the image embedding $\vx_T$. (We consider an image generation model here and discuss the extension to text-to-image models in \sectionautorefname~\ref{subsec:diss_conditional}.) Given a query image $\vx_0$, the attacker first runs the forward pass to obtain the image embedding $\vx_T$. Note that this requires knowledge or an educated guess about the total number of steps $T$ and the scheduling of the scaling factor $\alpha_t$. This information is typically displayed on online APIs that allow flexible control of the generation process (e.g., as the "num\_inference\_steps" and "scheduler" parameters\footnote{Hugging Face Stable Diffusion API: \href{https://replicate.com/stability-ai/stable-diffusion}{https://replicate.com/stability-ai/stable-diffusion}}). The intermediate outputs can be obtained by controlling the number of inference steps $t$ and extracting the corresponding output images displayed on the API. The attack can be formulated as:
\begin{align}
\label{eq:grey_box_attack}
&\mathcal{A}(\vx^i,\gM(\vtheta)) = 
\mathds{1}\Big[f\Big(\big\{\widehat{\Ls_{t}}(\vtheta,\vx^i)\big\}_{t=0}^{T_\mathrm{trun}}\Big)<\tau(\vx^i)\Big] \\
& \text{with} \quad \widehat{\Ls_{t}}(\vtheta,\vx^i) = \Vert \hat{\vx}_\vtheta (\vx_t^i,t) - \vx_0^i\Vert \nonumber
_2^2 \end{align}
Compared to the white-box setting, the attack assumption is slightly relaxed in that the attacker needs to estimate the loss terms based on the information typically available on online APIs. Each term of $\widehat{\Ls_{t}}$ differs from the ground-truth $\Ls_{t}$ used in the white-box case by a scaling factor (see \equationautorefname~\ref{eq:predict_x}). We deliberately do not use this scaling factor to reduce the impact of the attacker not knowing the exact $\alpha_t$ (this is accessible in some existing APIs but not all of them) and potentially making incorrect guesses in some cases. Additionally, we use the same truncation trick and explore several statistic functions $f$ as in the white-box case to encourage distinguishability in the membership score. 

Additionally, we explore the situation where the model owner may reduce the intermediate outputs by subsampling the inference steps, for example, to speed up the generation or limit potential privacy exposure. Formally, the attack in this case can be formulated as follows:
\begin{align}
&\mathcal{A}(\vx^i,\gM(\vtheta)) = 
\mathds{1}\Big[f(\gS)<\tau(\vx^i)\Big] \\ 
& \text{with} \quad \gS \subseteq \big\{\widehat{\Ls_{t}}(\vtheta,\vx^i)\big\}_{t=0}^T \nonumber 
\end{align}
That is, the attacker may only have access to a subset of the intermediate outputs from the reverse denoising steps of the diffusion model. We delve into the truncation techniques specific to this scenario in the experiment section. 

\begin{table}[!t]
\centering
  \begin{adjustbox}{max width=\columnwidth}
  \begin{tabular}{lcccc}
  \toprule %
  &  Model  & Image & Model & Hyperparameters \\
  &   type  & embedding & internals & $T, \alpha_T$ \\
  \midrule
  white-box (\ref{subsec:whitebox}) & $\checkmark$ &  $\checkmark$ & $\square$  & $\checkmark$ \\
  \hdashline
  grey-box (\ref{subsec:greybox}) &  $\checkmark$ &  $\checkmark$ & $\blacksquare$  & $\checkmark$ \\
  grey-box extension (\ref{subsec:diss_conditional}) &   $\checkmark$ &  $\times$ & $\blacksquare$  & $\checkmark$ \\
  \hdashline 
  black-box specific (\ref{subsec:blackbox}) &  $\checkmark$ &  $\times$ & $\blacksquare$ & $\times$ \\
  black-box agnostic (\ref{subsec:blackbox}) & $\times$ & $\times$ & $\blacksquare$ & $\times$ \\
  \bottomrule
  \end{tabular}
  \end{adjustbox}
  \caption {Taxonomy of attack settings. ($\times$: without access; $\checkmark$: with access; $\blacksquare$: black-box; $\square$: white-box). "Model type": Whether the underlying model is known to be a diffusion model. "Image embedding": Whether access to the image embedding during generation is available.
  }
  \label{table:taxonomy}
 \end{table}
 
\subsection{Black-box setting}
\label{subsec:blackbox}
\myparagraph{Threat Model}
In this scenario, attackers are limited to passively obtaining generated samples from well-trained generative models, without the ability to affect the generation process.  This creates a realistic scenario as there are no set assumptions about the attacker's abilities. We categorize the situation into two cases based on the attacker's knowledge of the synthetic data being generated by a diffusion model, referred to as "\textit{known model type}" and "\textit{unknown model type}". In the "\textit{known model type}" case, the attacker recognizes that the accessible synthetic data was produced by a diffusion model and may exploit this information to design targeted attacks. For instance, this may correspond to common situations where the attacker can only collect final outputs from online diffusion model APIs but is not allowed to perform steerable generation, thereby preventing our gray-box attacks discussed in \sectionautorefname~\ref{subsec:greybox}.

\myparagraph{Approach}
\textit{Model-specific Attack}: Once the attacker knows that the synthetic data set was generated by a diffusion model, a natural approach would be to train a shadow diffusion model to imitate the target diffusion model by using the synthetic data set as the training set. This enables the attacker to carry out an attack in the same way as in a white-box scenario. This type of attack is referred to as a \textit{"model-specific attack"} to differentiate it from attacks that do not use any information about the generative models. Formally, this can be expressed as:
\begin{equation}
\label{eq:blackbox_shadow}
\mathcal{A}(\vx^i,\gM(\vtheta)) = 
\mathds{1}\Big[f\Big( \big\{ \Ls_{t}(\vtheta^s,\vx^i)\big\}_{t=0}^{T_\mathrm{trun}}\Big)<\tau(\vx^i)\Big] 
\end{equation}
where $\vtheta^s$ represents the parameters of the shadow model, which were obtained by training the diffusion model on the synthetic data generated by the target model (parameterized by $\vtheta$).

\textit{Model-agnostic Attack}: In the absence of any additional information except for the synthetic sample set, the attacker's last resort is to use \textit{model-agnostic attacks}. Several options exist in the literature, such as GAN-Leak~\cite{chen2020gan}, which uses the Euclidean distance to the nearest neighbor in the synthetic set as a proxy for the sample loss and membership score, and Monte-Carlo~\cite{hilprecht2019monte}, which counts the number of generated samples within an $\epsilon$-ball of the query using a carefully designed distance metric. In line with previous work, we use the distance of the query image to its nearest neighbor in the synthetic set as the membership score. Furthermore, we enhance the distance metric by using a pre-trained feature extractor (trained on the large-scale public ImageNet~\cite{deng2009imagenet} dataset) and further refine the distance calculation by leveraging label information if available.
Formally,
\begin{equation}
\label{eq:blackbox_aganostic}
\mathcal{A}(\vx^i,\gM(\vtheta)) = 
\mathds{1}\Big[\min_k \, \Big\{ \ell_{dis}(\vx^i, \vs^k ) \Big\}_{k=1}^K 
<\tau(\vx^i)\Big]
\end{equation}
with $\vs^k \sim p_\vtheta$ representing the samples generated by the target model parameterized by $\vtheta$. $K$ denotes the total number of synthetic samples, and $\ell_{dis}$ is the cosine distance in the feature space of a pre-trained ImageNet classifier, where the feature space is determined by the output of the second last layer.

\subsection{Analysis \& Insights}
\label{subsec:analytical_insights}
While our attack and several previous ones fall within a general likelihood ratio formulation (\equationautorefname~\ref{eq:opt_ratio} in \sectionautorefname~\ref{sec:background_MI}), and therefore demonstrate analogous algorithmic components, our attack  possesses distinctive features that are critical to its superior effectiveness. Specifically, 
the white-box attack in \cite{hilprecht2019monte} uses reconstruction error as the membership score. Adapting it to diffusion models results in a special case of our gray-box attack (using the loss term at the final time step), but with suboptimal configurations, i.e., truncating all steps except the final one or using the ``Min'' statistical function (see results in \sectionautorefname~\ref{subsec:eval_grey_box}).  
The adaptation of the GAN-Leaks~\cite{chen2020gan} white-box attack to diffusion models would suggest optimizing the latent code using a gradient-based method to find the nearest neighbor in the output space of the target diffusion model and using the nearest-neighbor distance as the membership score. This approach will be upper-bounded by the aforementioned white-box attack of \cite{hilprecht2019monte} (that does not require optimization over the latent space), and thus further upper-bounded by our methods.  
Seen from a more abstract perspective, the GAN-Leaks framework (suggests using the approximated data-likelihood) would translate into our white-box attack with a ``Sum'' statistical function, which is also a suboptimal special case of ours.  
Furthermore, while previous black-box model-agnostic attacks are applicable to diffusion models, we show in Sections~\ref{subsec:eval_blackbox} that our attack generally outperforms previous ones (Figures~\ref{fig:blackbox_celeba_comparison}-\ref{fig:blackbox_cifar_comparison}).  

The primary factors motivating the specialized features of our attack design lie in the intrinsic properties of diffusion models. 
Firstly, diffusion models generate data iteratively, a characteristic that sets them apart from most other types of generative models. This formulation offers key opportunities for the attack to exploit knowledge from multiple iterations, instead of relying solely on the singular one-shot signal as done in previous attacks. Moreover, the loss terms produced from different iterations may convey varying amounts of information. This variability necessitates specialized treatment, such as the truncation technique used in our method.
Specifically, by construction of diffusion models, the Signal-to-Noise Ratio (SNR)
for the latent variable distribution at each time step conditioned on the clean sample $q(\vx_t|\vx_0)$, decreases monotonically as the time step increases~\cite{kingma2021variational}. In other words, as $q(\vx_t|\vx_0)$$=$$ \gN(\vx_t;\sqrt{\bar{\alpha}_t}\vx_0, (1-\bar{\alpha}_t)\mI)$ (\equationautorefname~\ref{eq:q_xt}), $\mathrm{SNR}(t)$$=$$\frac{\mu^2}{\sigma^2}$$=$$\frac{\bar{\alpha}_t}{1-\bar{\alpha}_t}$, with $\mathrm{SNR}(t)$$\,<\,$$\mathrm{SNR}(s)$ for all $t\,$$>$$\,s$.
This inherent property results in loss terms that are closer to the Gaussian noise end becoming larger, since the latent variable tends to carry less informative signals about the samples $\vx_0$ as $t$ increases (as illustrated in \figureautorefname~\ref{fig:trajactory}).
Our attack design  accommodates these unequal weights for each term by permitting arbitrary statistical functions that can
capture the distinctive characteristics for various scenarios. Additionally, our method explicitly eliminates terms that, while dominant in scale, provide less discriminative information.

% In contrast to Generative Adversarial Networks (GANs), which belong to implicit density models without an encoding process during training, and Variational Autoencoders (VAEs), which learn the encoding process across general data distributions, diffusion models employ a predetermined (unlearnable) encoding process. This allows a precise association between the sample image and its latent variance, thus reducing the attack's effort in acquiring latent codes in diffusion models. Consequently, we allocate less focus to this step during attack design.

\section{EXPERIMENT ANALYSIS}
\label{sec:experiments}
%\mario{i'm sorry that I've missed this point: is there no proof of concept on a large diffusion model? ideally taking one that is available pretrained - and showing MIA performance or some annecdodale evidence on this? i'm not sure if reviewers can be convinced by CelebA or CIFAR-10 if it comes to diffusion models. a reviewer might argue that models trained on large datasets behave very differently. is there any way we can argue about this?}
%-------------------------------------------------------------------------------
In this section, we present the first systematic evaluation of membership inference attacks against state-of-the-art diffusion models. Our comprehensive study encompasses all viable threat models, ranging from the most knowledgeable white-box setting to the most practical black-box one. Importantly, we make key discoveries linking the attack effectiveness to the threat model, data set, model architecture, and training configuration, leading to practical implications for securing the deployment of diffusion models in real-world settings.

\subsection{Setup}
\myparagraph{Datasets} In line with previous research on MIAs against generative models~\cite{hayes2019logan,chen2020gan}, we conduct experiments on benchmark image datasets with diverse characteristics:
\begin{itemize}
    \item \textit{CelebA~\cite{liu2015faceattributes}} is a large-scale face attributes dataset containing 200k RGB images, which are aligned using facial landmarks. To ensure comparability with previous results, we adopt the standard pre-processing procedure when training diffusion models and evaluating attack performance. This involves randomly selecting a maximum of 20k images (corresponding to the more challenging random-split setting in~\cite{chen2020gan}), center-cropping them, and resizing them to a resolution of $64$$\times$$64$ for training the models and evaluating the attacks.
    \item \emph{CIFAR-10~\cite{krizhevsky2009learning}}
 is a dataset of 60k RGB images with shape $32$$\times$32$\times$$3$. Each image is labeled with one of 10 classes, representing the object depicted in the image.
\end{itemize}
In every setup, we assess the attack effectiveness on a balanced query set $\sS$, i.e, with 
an equal number of members and non-members.

\myparagraph{Target Models}
We evaluate state-of-the-art diffusion models using their official PyTorch implementation: \textbf{Improved Diffusion}~\cite{nichol2021improved}\footnote{\href{https://github.com/openai/improved-diffusion}{https://github.com/openai/improved-diffusion}}, \textbf{Guided Diffusion}~\cite{dhariwal2021diffusion} \footnote{\href{https://github.com/openai/guided-diffusion}{https://github.com/openai/guided-diffusion}}  and pre-trained \textbf{Stable Diffusion}~\cite{Rombach_2022_CVPR}\footnote{\href{https://huggingface.co/CompVis}{https://huggingface.co/CompVis}} models (see the Appendix for more details). By default, we set the number of denoising steps $T$ to be 4000 and adopt a standard linear scheduler for $\alpha_t$. We mainly focus on the unconditional image generation task and investigate text-to-image generation model in \sectionautorefname~\ref{subsec:diss_conditional}. All experiments were conducted on a single NVIDIA A100 GPU. \\
It is essential to ensure the generation quality of the target models, as attackers are more likely to target models with high utility and practical significance. We present both qualitative and quantitative evaluations of the generation quality in terms of the Fr\'{e}chet Inception Distance (FID) metric~\cite{heusel2017gans} (as shown in \tableautorefname~\ref{tab:comparison_gan_diffusion} and \figureautorefname~\ref{fig:imgs_samples_from_different_generative_models}). A smaller FID value indicates that the generated images are more realistic and closer to the distribution of real data. In particular, our target models generate high-quality output, surpassing the results in previous works~\cite{hayes2019logan,hilprecht2019monte,chen2020gan}, demonstrating the high practical value of our study.

% Generative Adversarial Networks (GANs) and Diffusion-based models are two popular types of generative models used for image synthesis and data generation. We compare the privacy risks of these two types of generative models. To demonstrate the serious privacy vulnerability in diffusion models, we compare the MIA attacks between the acknowledged simple GAN model, PGGAN and two diffusion-based models, Guided-diffusion and Improved-diffusion models.

\myparagraph{Attack Evaluation} The proposed membership inference attack is formulated as a binary classification as described in Equations~\ref{eq:white_box_attack}--\ref{eq:blackbox_aganostic}, with a threshold $\tau$. For simplicity, a sample-independent threshold is used.
The attack performance is evaluated by measuring the area under the receiver operating characteristic curve (\textbf{AUCROC}), which is obtained by varying $\tau$. The complete \textbf{ROC curve} is also provided for clear visualization of the attack's properties~\cite{carlini2022membership}. Moreover, we evaluate the truth false positive rate under a certain low false positive rate (i.e., \textbf{TPR@1\%FPR} and \textbf{TPR@0.1\%FPR}) to demonstrate the attack performance with realistic scenarios~\cite{carlini2022membership}. Additionally, following \cite{hayes2019logan}, the attack \textbf{Accuracy} and \textbf{F1 Score} are calculated by setting the threshold to the median value of the membership scores over the query set. All these metrics have a value ranging from 0 to 1, with a higher value indicating a more effective attack.

\subsection{Evaluation on White-box attack}
\label{subsec:eval_white_box}

\begin{figure}[!t]
\centering
\includegraphics[width=0.9\columnwidth]{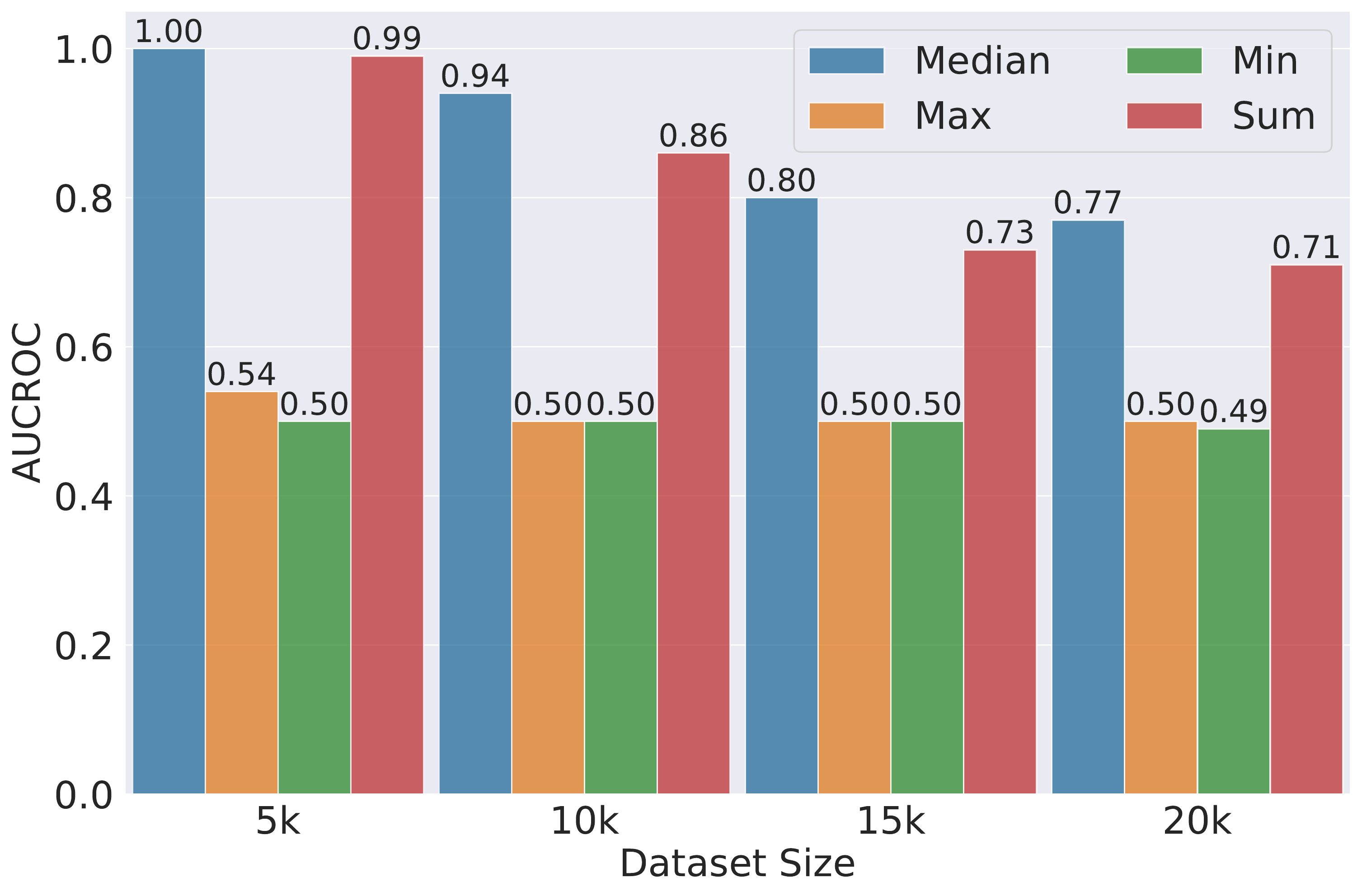}
\caption{The \textbf{white-box} attack AUCROC when applying different statistic function $f$ (\textit{Sum}, \textit{Median}, \textit{Min}, and \textit{Max}) to the \textit{entire loss trajectory} $\{\Ls_{t}\}_{t=0}^T$ on CelebA. The \textit{``Sum''} function corresponds to the direct use of  $\Ls_{vlb}$ for MIA. }
\label{fig:whitebox_statistical_approaches_comparison_vb}
\end{figure}

\begin{figure*}[!t]
\centering
\subfigure[White-box]{
\includegraphics[width=0.3\textwidth]{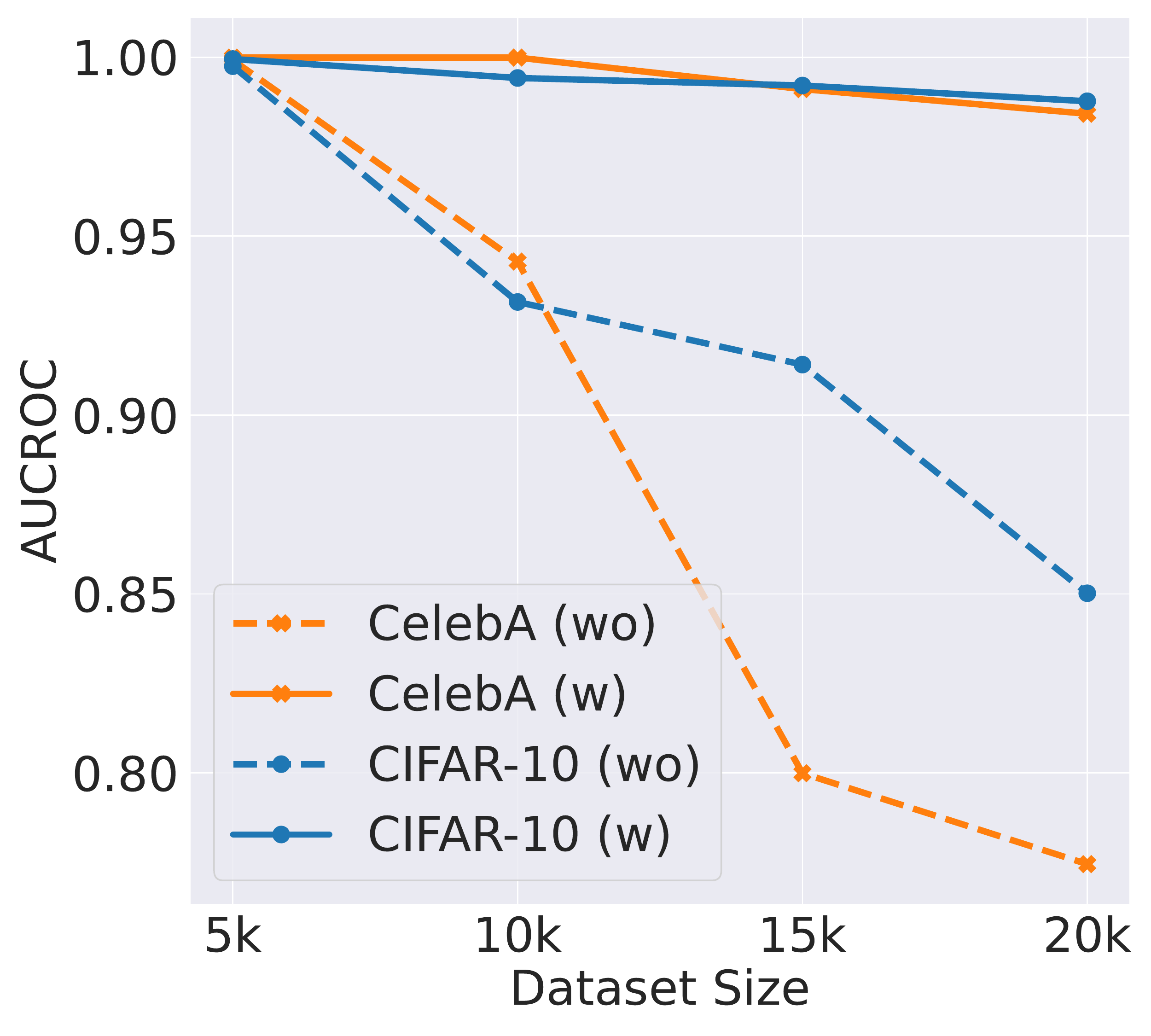}
\label{fig:wb_size}
}
\subfigure[Gray-box]{
\includegraphics[width=0.3\textwidth]{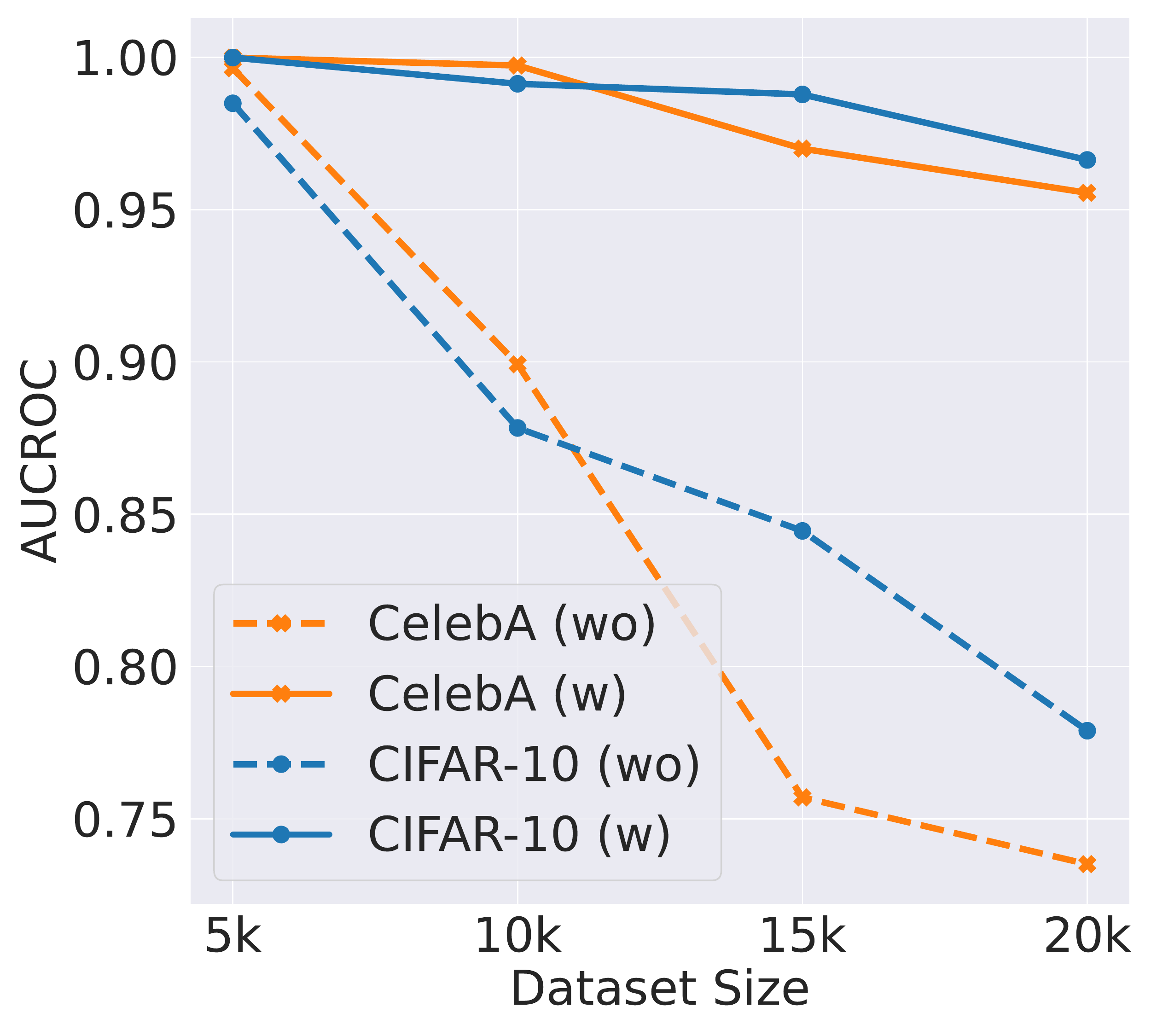}
\label{fig:gb_size}
}
\subfigure[Black-box (model-specific)]{
\includegraphics[width=0.3\textwidth]{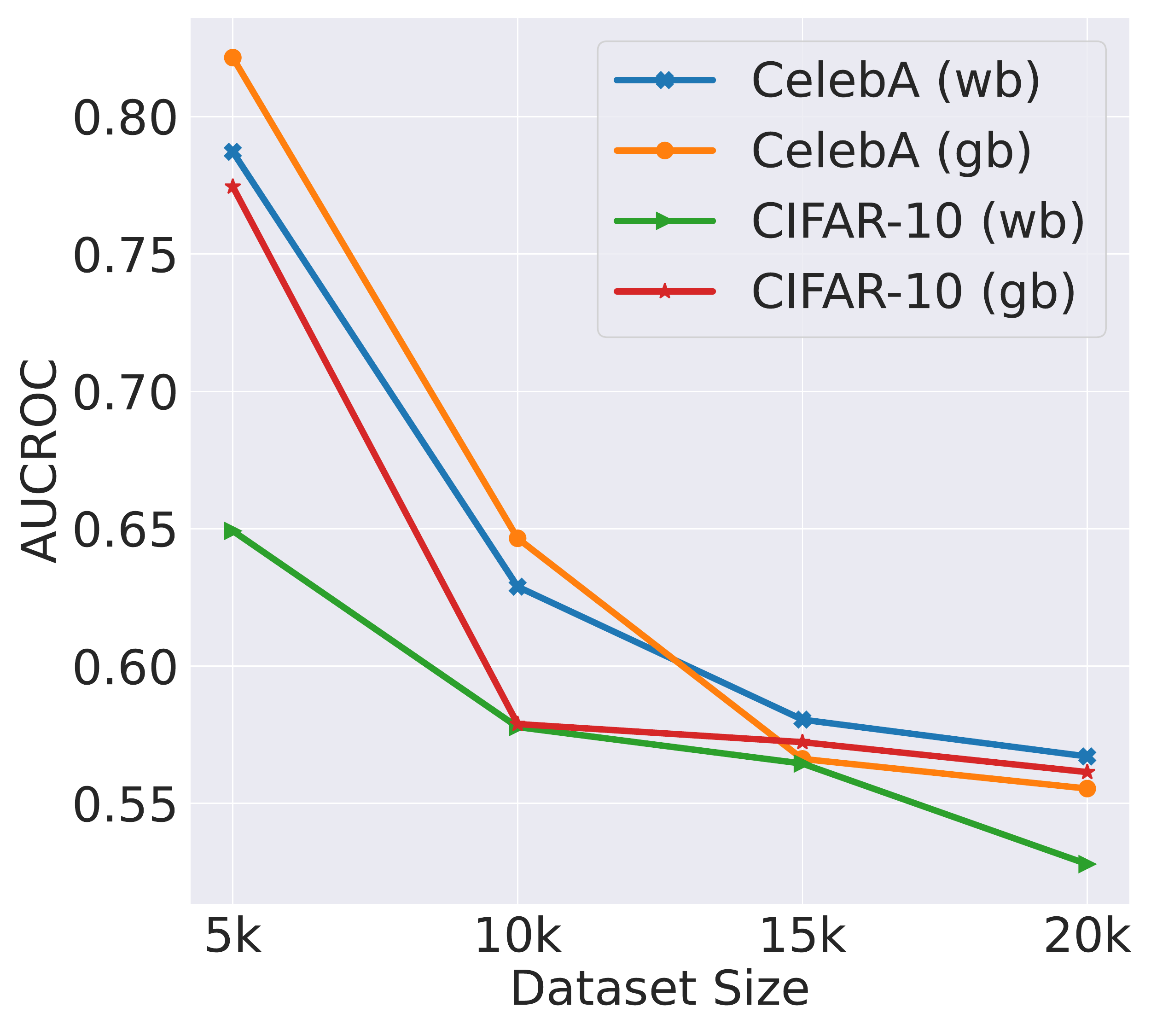}
\label{fig:bb_size}
}
\caption{The attack AUCROC across different dataset sizes and attack scenarios. The results obtained with (indicated as ``w'' and shown as solid lines) and without (indicated as ``wo'' and shown as dashed lines) applying our truncation techniques are compared. We present the best results for each case and the truncation step is always set to be $0.75T$.
}
%\caption{The attack AUCROC across different dataset sizes and attack scenarios. The results obtained with (indicated as "w" and shown as solid lines) and without (indicated as "wo" and shown as dashed lines) applying our truncation techniques are compared. We present the best results for each case where the \textit{Median} function is used for "wo" truncation and \textit{Max} function is adopted for "w" truncation, the truncation step is always set to be $0.75T$.}
\end{figure*}
\myparagraph{Effectiveness of Sample Losses as Membership Score}
We first assess the feasibility of inferring membership with white-box access to a target diffusion model based on the sample loss terms $\{\Ls_t\}_{t=0}^T$ and/or the VLB loss $\Ls_{vlb}$ (which represents the sum of all loss terms). As shown in Figure~\ref{fig:whitebox_statistical_approaches_comparison_vb}, even simply using $\Ls_{vlb}$ as the membership score achieves promising attack performance. For instance, we obtain 0.62 AUCROC when using $\Ls_{vlb}$ to attack the target model trained with 20k data samples. The performance improves to 0.68 when we explore various options of the statistical function $f\in$ \{\textit{Sum, Median, Max, Min}\} and select the best, which is the ``\textit{Median}'' in this setting. As a reference, previous work reported a maximum AUCROC of 0.61 
 under comparable conditions (i.e., white-box attack with the same size and type of split in the training set, without using additional reference data) when attacking GANs~\cite{chen2020gan}.  The ``\textit{Average}'' function is not considered as it is equivalent to ``\textit{Sum}'' in terms of discrimination. While we expect that a more complicated design of function $f$ might lead to improved results, we observe that a simple data-independent function works sufficiently well in most cases and stick to such choices throughout our evaluation.

\myparagraph{Performance Gain from Truncation}
To address the instability and indistinguishability caused by large variations in the magnitude of $\Ls_{t}$ (as discussed in \sectionautorefname~\ref{subsec:analytical_insights} and depicted in \figureautorefname~\ref{fig:whitebox_vb_trajactory}), we improve our attacks by truncating the loss trajectory. This involves excluding the initial denoising steps that have limited relevance to membership but have high values that can easily dominate the statistical function. We present the detailed results in \tableautorefname~\ref{tab:wb_celeba_truncation_all} (in Appendix) and present the results for the best configuration ($f$ is selected to be ``\textit{Max}'' and $T_{trun}$ is set to be $0.75T$) in \figureautorefname~\ref{fig:wb_size}. As shown in \figureautorefname~\ref{fig:wb_size}, our truncation techniques consistently improve the attack performance across various training configurations. In the most challenging setting with a dataset size larger than 15k, the improvement is particularly significant: by around 0.2 AUCROC on  CIFAR-10 and 0.25 on CelebA, respectively.

We further show that the performance gain is \textit{not} sensitive to the particular choice of the truncation step $T_{trun}$ and training configurations. As can be seen from~\tableautorefname~\ref{tab:whitebox_truncate_steps_exploration} (See \figureautorefname~\ref{fig:white_vlb_truncation} for the visualization), a boost in the attack performance can be achieved with a relatively large range of possible values of $T_{trun}$ (e.g., when $T_{trun}$ is roughly in the range from $0.875T$ to $0.75T$). Moreover, the best value turns out to be consistent across different training settings (i.e., the training set size and dataset in our experiments). This suggests a high practical value of our technique such that the attacker may be able to determine the appropriate parameter on any available reference dataset and use such selected parameters for completing the attack. We set by default the statistic function to be a ``\textit{Max}'' function and $T_{trun}$ to be $0.75T$ for our white-box attack when adopting truncation techniques, which empirically leads to promising performance across various situations.

\begin{table}[!ht]
\aboverulesep=0ex
\belowrulesep=0ex
\centering
\definecolor{Gray}{gray}{0.9}
\newcommand{\cc}{\cellcolor{Gray}}
\resizebox{\columnwidth}{!}{%
\begin{tabular}{c|ccccccc}
\toprule
Size & ref & \cc $T$ & $0.975T$ & $0.875T$ & $0.75T$ & $0.625T$ & $0.5T$ \\
 \midrule
 5k & 1.00 & \cc 0.54 & \textbf{1.00}	& \textbf{1.00}	& \textbf{1.00}	& \textbf{1.00}	& \textbf{1.00}	\\
 10k  & 0.94 & \cc 0.50 & 0.97	& \textbf{1.00}	& \textbf{1.00}	& 0.99	&0.93	\\ 
 15k & 0.80 & \cc 0.50 & 0.81	& \textbf{0.99}	& \textbf{0.99}	& 0.96	&0.80	\\ 
 20k & 0.77 & \cc 0.50 & 0.65	& 0.97	& \textbf{0.98} & 0.95	& 0.77	\\ 
 \bottomrule
%\midrule
%5k & 1.00 & \cc 0.54	& \textbf{1.00}	&\textbf{1.00}	&\textbf{1.00}	&\textbf{1.00}	& \textbf{1.00}\\
%10k  & 0.91 & \cc 0.50	& 0.97	& \textbf{1.00}	& \textbf{1.00}	&0.99	&0.91 \\ 
%15k & 0.76& \cc 0.50	& 0.76	&0.97	&\textbf{0.98}	&0.94	&0.76\\ 
%20k & 0.68 & \cc 0.49	& 0.65	& 0.91 &\textbf{0.93}	& 0.86	& 0.68\\ 
%\bottomrule
\end{tabular}
}
\caption{The \textbf{white-box} Attack AUCROC for different truncating steps $T_{trun}$ in CelebA trained with different training data size. The \textbf{``ref''} (meaning ``reference'') column represents the best results that can be achieved without any truncation (but may correspond to different statistic functions). The rest of the results correspond to using a ``\textit{Max}'' statistic function. The columns, from left to right, represent increasing levels of truncation, excluding the top $[0, 100, 500, 1000, 1500, 2000]$ steps, respectively. The column labeled ``$T$'' (in gray) represents ``no truncation''. The best results (selected to four decimal places) for each configuration are highlighted in bold.}
\label{tab:whitebox_truncate_steps_exploration}
\end{table}

\myparagraph{Effect of Dataset Size} The size of the training dataset is a key determinant of the membership risk associated with machine learning models, as previously noted in the literature~\cite{shokri2017membership,hayes2019logan,chen2020gan}. As the number of training samples grows, the model becomes unable to capture the point-wise delta distribution and moves from memorization to generalization. 
Specifically, previous studies have shown that MIA performance tends to be less effective ($<0.6$ AUCROC) when the training dataset size exceeds 10k~\cite{hilprecht2019monte,chen2020gan}. 

Consistent with these findings, our results show a tendency of decline in the attack AUCROC as the size of the training dataset grows (refer to Figure~\ref{fig:wb_size}). However, we observe a consistently high level of attack performance throughout our evaluations, even in cases where previous attacks have typically failed. For example, when the training set size is up to 10k, our attack achieves near-perfect AUCROC on both datasets, and even with a training set size of 20k, our attack remains highly effective (0.98 attack AUCROC for CelebA and 0.99 for CIFAR-10) after applying our truncation techniques. %Additionally, even when the training set size reaches 20k, our attack still achieves an exceptional performance (AUCROC $>0.95$). 
These results highlight the potential of using sample losses for effective attacks and the significant privacy risk incurred by the common practice of sharing diffusion models in open source.

%Previous research has disclosed that dataset size is highly related to the vulnerability of GAN, a popular kind of generative model, to MIA~\cite{chen2020gan}. In general, a larger dataset led to less training samples memorization in models. Therefore, many previous works show that the MIA failed in GAN when the dataset size increased to larger than 10,000~\cite{hilprecht2019monte,chen2020gan}. As the diffusion models are considered as an alternative to GAN in image(text) generations. It is important to explore how the dataset size affects the vulnerability to MIA in diffusion models.

\begin{comment}
\begin{table}[]
\resizebox{\columnwidth}{!}{%
\begin{tabular}{c|cccc}
\hline
Size of Training Set & 5000 & 10000 & 15000 & 20000 \\ \hline
Total VLB             & 0.93 & 0.69  & 0.59  & 0.58  \\ \hline
\textit{DiffMIA}-VB(Median)    & 0.99 & 0.97  & 0.91  & 0.88  \\ \hline
\end{tabular}%
}
\caption{The attack AUC with total VLB score and truncated VB score with different sizes of training data in the Cifar10 dataset.}
\label{tab:vb_total_bpd_comparsion}
\end{table}
\end{comment}

\subsection{Evaluation on Gray-box attack}
\label{subsec:eval_grey_box}

\myparagraph{Estimated Losses as Membership Score} 
We further consider the real-world scenario where third-party providers, such as Amazon AWS, offer API services to create images using diffusion models. In this scenario, attackers typically have access to the generated images at any inference step of the deployed diffusion model (e.g., by specifying the inference step parameter and obtaining the displayed output), but may not have knowledge of the ground-truth loss terms $\Ls_t$ (which requires knowing the exact values of $\alpha_t$). As discussed in Section~\ref{subsec:greybox}, the attack must estimate the loss based on the intermediate output shown on the API. We evaluate the attack performance based on our proposed estimation in \equationautorefname~\ref{eq:grey_box_attack}. 

\begin{figure}[!t]
\centering
\includegraphics[width=0.9\columnwidth]{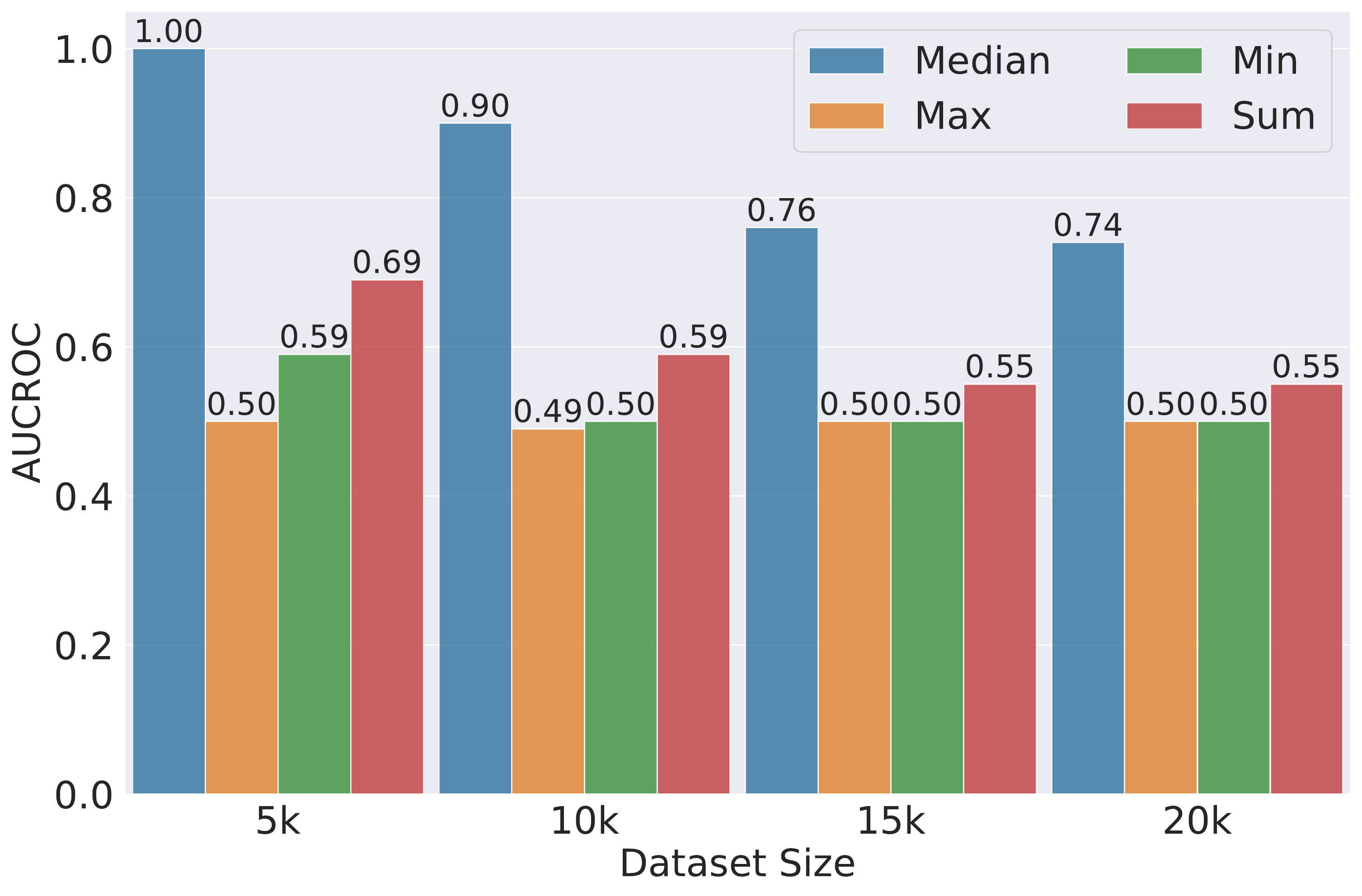}
\caption{The \textbf{gray-box} attack AUCROC when applying different statistic function $f$ (\textit{Sum}, \textit{Median}, \textit{Min}, and \textit{Max}) to the entire loss trajectory estimated based on the intermediate outputs (i.e., $\{\widehat{\Ls}_{t}\}_{t=0}^T$) on CelebA.}
\label{fig:graybox_statistical_approaches_comparison_xstart_mse}
\end{figure}

We present our results obtained by using the whole estimated loss trajectories $\{\widehat{\Ls}_t\}_{t=0}^T$ in \figureautorefname~\ref{fig:graybox_statistical_approaches_comparison_xstart_mse}. As can be seen, we demonstrate promising performance with an AUCROC of $0.9$ for datasets with up to 10k training samples, and an AUCROC value of $0.74$ when the dataset size grows to 20k.
Despite a slight decrease in comparison to the white-box setting, our gray-box already achieves a reasonable level of performance in its vanilla form, suggesting the potential effectiveness of our formulation. Our results also reveal that ``\textit{Median}'' and ``\textit{Sum}'' statistical functions perform better than the others, with ``\textit{Median}'' showing the best performance in most cases across different training configurations. This naturally follows our intuition that using a robust statistic that captures the discriminative factors may be preferable over simply aggregating the available information. Moreover, when compared to the white-box case, the loss terms used by our gray-box method generally exhibit higher variance and magnitude (caused by the difference in the scaling factors between the white-box and gray-box loss terms). This observation partially explains the superior performance of different functions in each scenario. Specifically, ``Median'' function is a more robust choice for the gray-box attack, while the ``Max'' function is a more discriminatory choice for the white-box attack.

\begin{comment}
\begin{table}[!ht]
\begin{adjustbox}{max width=\columnwidth}
\definecolor{Gray}{gray}{0.9}
\newcommand{\cc}{\cellcolor{Gray}}
\aboverulesep=0ex
\belowrulesep=0ex
\centering
\begin{tabular}{c|cccc}
\toprule
Size &  Min & Max & Median & Sum \\
\midrule
5k  &  \\ 
10k &   \\ 
15k  &  \\
20k &  \\ 
\bottomrule
\end{tabular}%
\end{adjustbox}
    \caption{The gray-box attack AUCROC when applying different statistic function $f$ (\textit{Min}, \textit{Max}, \textit{Median}, and \textit{Sum}) to the \textit{entire loss trajectory estimated based on the intermediate outputs} $\{\widehat{\Ls}_{t}\}_{t=0}^T$. The experiments were conducted on the CelebA dataset with various training set sizes, as indicated in the first column \dingfan{(also draw the figure).} }
\label{tab:graybox_whole_trajectory_celeba}
\end{table}
\end{comment}

% change the figure to generation process figures for member , nonmember and boardline
\begin{figure*} [!h]
	\centering
 \hspace*{0.2cm}\includegraphics[width=0.99\textwidth]{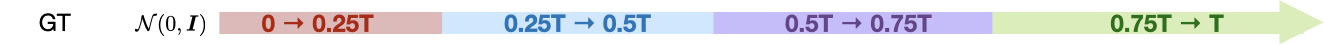}\vspace{-4pt}
	\subfigure[Member samples (correctly identified)]{
 \includegraphics[width=\textwidth]{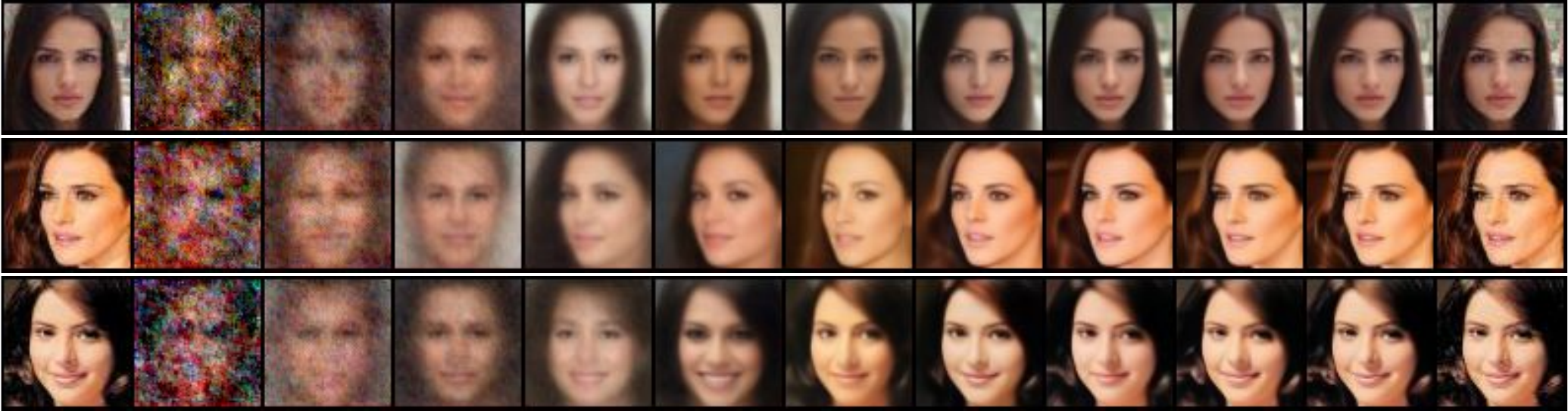}
		\label{fig:member_generation_process}}
	\subfigure[Non-member samples (correctly classified)]{
		\includegraphics[width=\textwidth]{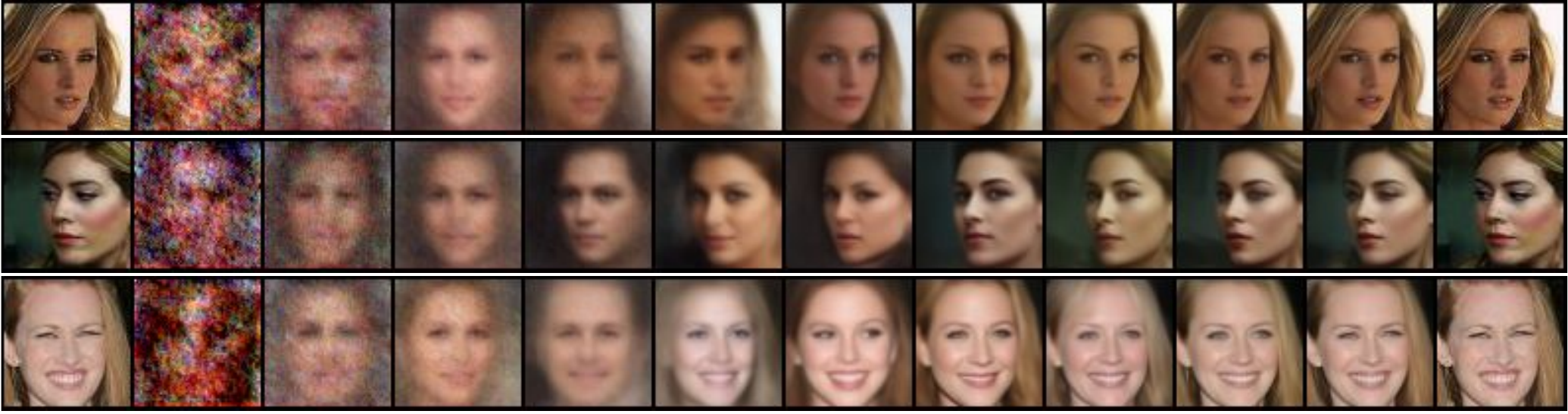}
		\label{fig:nonmember_generation_process}}
	 \subfigure[Member samples (but classified as non-members)]{ 
  	\includegraphics[width=\textwidth]{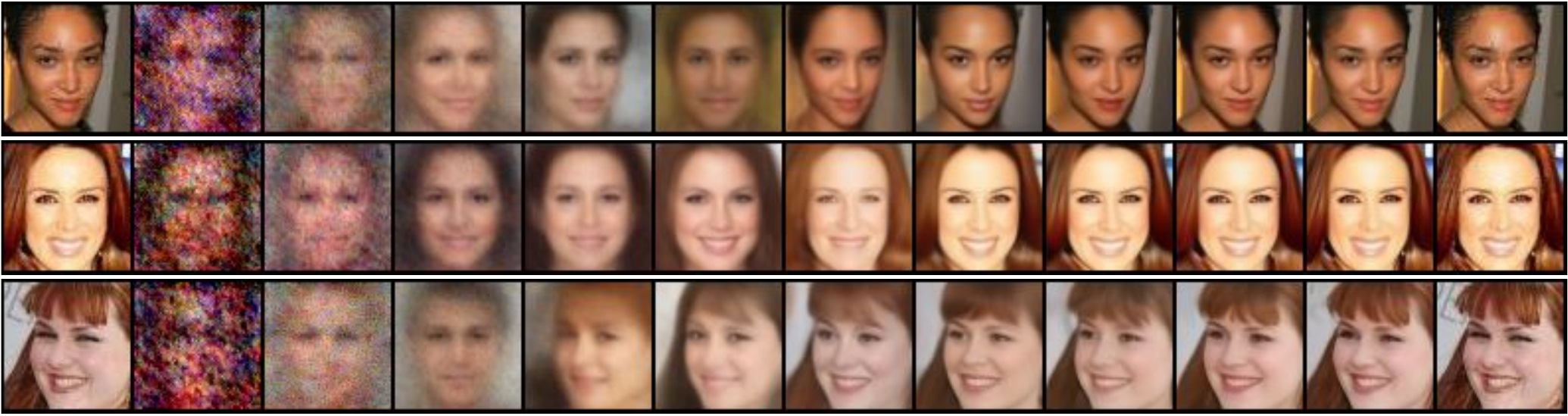}
	 	\label{fig:boardline_generation_process}}
	\caption{The query images together with their reconstruction triggered by our gray-box attack. The first column presents the ground-truth query images, while the last column refers to the final generated images from Guided-Diffusion models trained on CelebA (20k). The intermediate results are plotted per 400 time steps. }
\label{fig:intermediate_generation_process_guided_diffusion}
\vspace{-10pt}
\end{figure*}

\myparagraph{Performance Gain from Truncation} 
Similar to the case in the white-box setting, not all of the estimated loss terms are informative in distinguishing between members and non-members. Moreover, recall that the generation process in a diffusion model is designed to mimic the reverse denoising process, resulting in noisier outputs in the early denoising stages and thus a larger difference when compared with the clean query sample. By construction, this makes these loss terms corresponding to the earlier denoising steps to have larger magnitudes and to possibly dominate the attack prediction. 

Our truncation technique is an effective solution for addressing this issue in the gray-box setting. As demonstrated in \figureautorefname~\ref{fig:gb_size} and detailed in \tableautorefname~\ref{tab:gb_celeba_truncation_all} (in the Appendix), our truncation technique consistently improves the attack AUCROC by up to 0.22 on CelebA and 0.2 on CIFAR-10, particularly in challenging cases where the training set size is larger than 10k. These cases typically result in less successful attacks with AUCROC less than 0.6 for existing works~\cite{chen2020gan}, whereas we achieve highly effective attacks with AUCROC around 0.95 throughout our evaluation.

We further validate our intuition (discussed in \sectionautorefname~\ref{subsec:analytical_insights}) via qualitative results in \figureautorefname~\ref{fig:intermediate_generation_process_guided_diffusion}, where we generally observed that: (1) Starting from the noise end, the intermediate results from the first $0.25T$ steps do not manifest significant visual distinction, which supports our truncation technique to eliminate or reduce the influence of less informative terms. (2) For member data, the intermediate results begin to visually resemble the query at around the $0.5T$ time step counted from the noise end. In contrast, non-member images require more steps (approximately up to $0.75T$) to reach a similar level of visual similarity. This discrepancy suggests that the target diffusion model indeed displays distinct behaviors for member vs. non-member images, which can be exploited by an adversary. (3) Both member and non-member images can be reconstructed to a high degree of visual similarity by the final step. This observation clarifies why relying solely on the final reconstruction difference (as suggested by prior works~\cite{hilprecht2019monte,chen2020gan}) may not yield optimal effectiveness. (4)  While some member samples might display complex visual patterns (and/or be underrepresented in the distribution) making them more challenging to reconstruct and thus harder for an MIA to detect (see examples in \figureautorefname~\ref{fig:boardline_generation_process}), they still tend to be successfully reconstructed at earlier time steps  compared to non-members. Consequently, it remains possible for a stronger attack (e.g., with carefully tuned hyperparameters) to detect these samples.

We also studied the impact of various truncation step options on the attack performance. As shown in \figureautorefname~\ref{fig:greybox_mse_truncation} (in the Appendix), the optimal choice remains largely stable across different training setups (see detailed quantitative results in \tableautorefname~\ref{tab:wb_celeba_truncation_all}). The results indicate that improvements can be achieved with a wide range of reasonable choices. Based on these findings, we set the default truncation step to be $0.75T$ and the ``\textit{Median}'' statistical function as the default for evaluating gray-box attacks with truncation techniques.

Other techniques such as re-weighting and re-scaling the loss terms may also help reduce the significant variation in loss term magnitude and lead to improved results if carefully tuned. However, we believe our truncation technique is advantageous for its simplicity and effectiveness.

\myparagraph{Adaptive Defenses} 
\label{subsec:adpative_defense} While some of the diffusion APIs expose all the relevant hyperparameters for generation (and potential attack) and allow controllable synthesis, model owners may decide to withhold certain information to protect commercial interests and preserve privacy, which creates extra challenges for attackers. We study the impact of withholding hyperparameters in diffusion model APIs from an adaptive defense perspective.

We take a more in-depth investigation into the case where the adopted scheduler of $\alpha_t$ is not accessible. While the official implementation supports two scheduler options, we evaluate our attack performance by using a different scheduler than the one used to train the target model. This simulates a worst-case scenario where the attacker guesses the hyperparameter incorrectly. Our results show that our gray-box attacks remain effective even when using a different scheduler (see \tableautorefname~\ref{tab:greybox_rescheduler}), with AUCROC values of 0.91 and 0.65 for datasets of 5k and 20k, respectively. These results suggest that even with a different scheduler, samples can still be mapped to descriptive embeddings in the latent space, revealing information for attack during the reverse generation process. Additionally, as there are only a few options for the scheduler and the forward process is largely the same or highly similar for most diffusion models, it is likely that the attacker can guess the correct scheduler. This implies that withholding the scheduler may not eliminate the privacy risk.

We also consider the case where the model owner may choose to suppress the intermediate outputs. 
As demonstrated in Table~\ref{tab:greybox_respace}, even when 75\% or 50\% of the intermediate outputs during the reverse generation process are suppressed, our attack remains highly effective. While such suppression reduces the amount of information leaked to the public, thus diminishing potential risks, we posit that a well-designed attack using appropriate statistical techniques can still be successful. This premise is supported by an examination of the loss term distribution: as shown in Figure~\ref{fig:whitebox_xstart_mse_trajactory}, an attack can exploit a certain range of the discriminative region. Even with substantial suppression, if the attacker can extract a subset of the intermediate outputs within such region, an inference attack can still be successfully executed.
%Table~\ref{tab:greybox_respace} shows the results of suppressing 75\% and 50\% of the intermediate outputs during the reverse generation process. Our attack is still highly effective in this scenario. Although suppressing the outputs reduces the information leaked to the public and reduces potential risk, we believe that the attacker may still be successful by using proper statistics. This can be seen by examining the distribution of the loss terms, as shown in Figure~\ref{fig:whitebox_xstart_mse_trajactory}. The attack can exploit a certain range of the discriminative region, and even with suppression, if the attacker can extract a subset of the intermediate outputs, the attack can still perform successful inference.

\begin{figure*}[!t]
\centering
\subfigure[CelebA]{
\includegraphics[width=0.9\columnwidth]{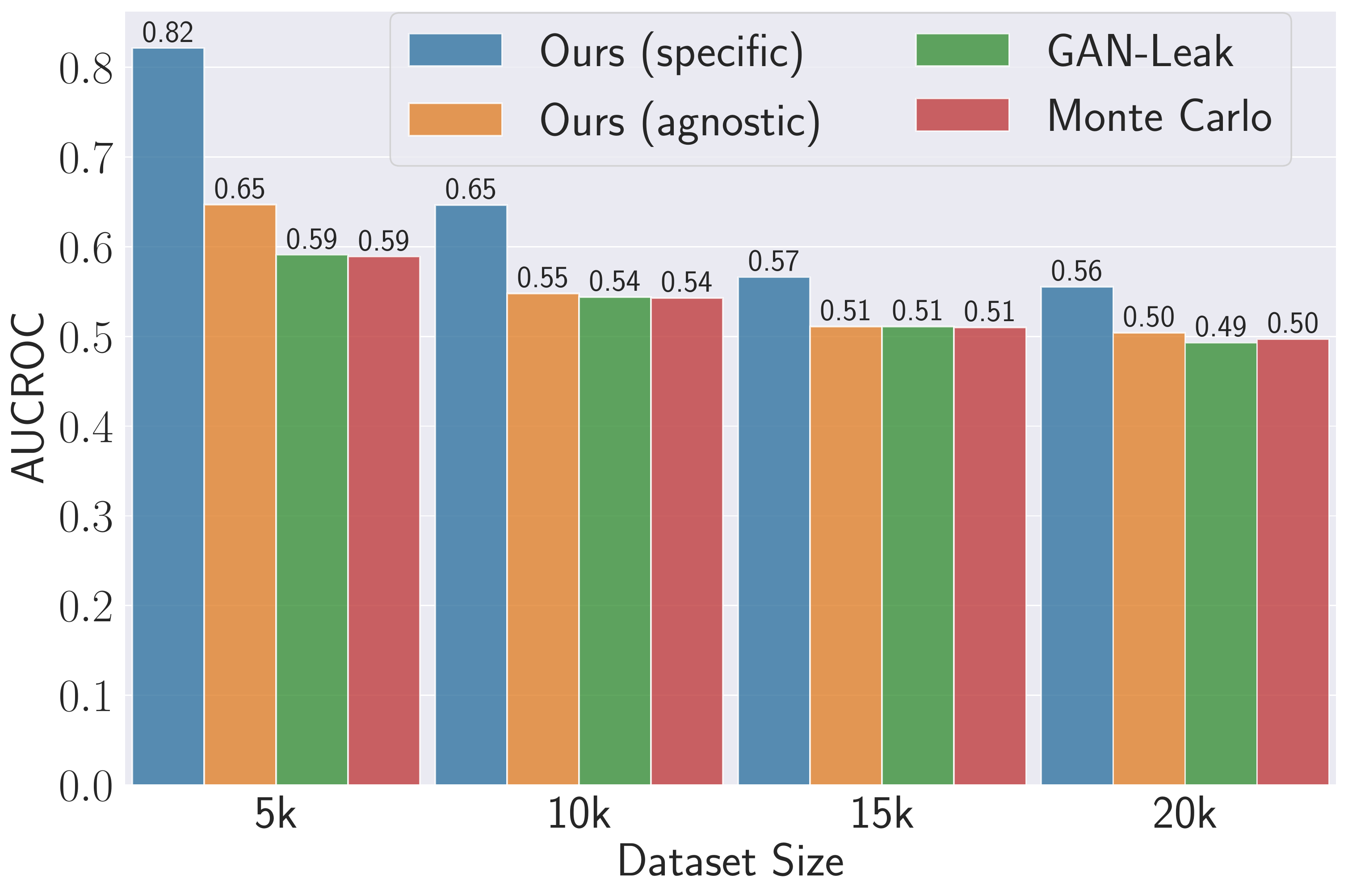}
\label{fig:blackbox_celeba_comparison}
}
\subfigure[CIFAR-10]{
\includegraphics[width=0.9\columnwidth]{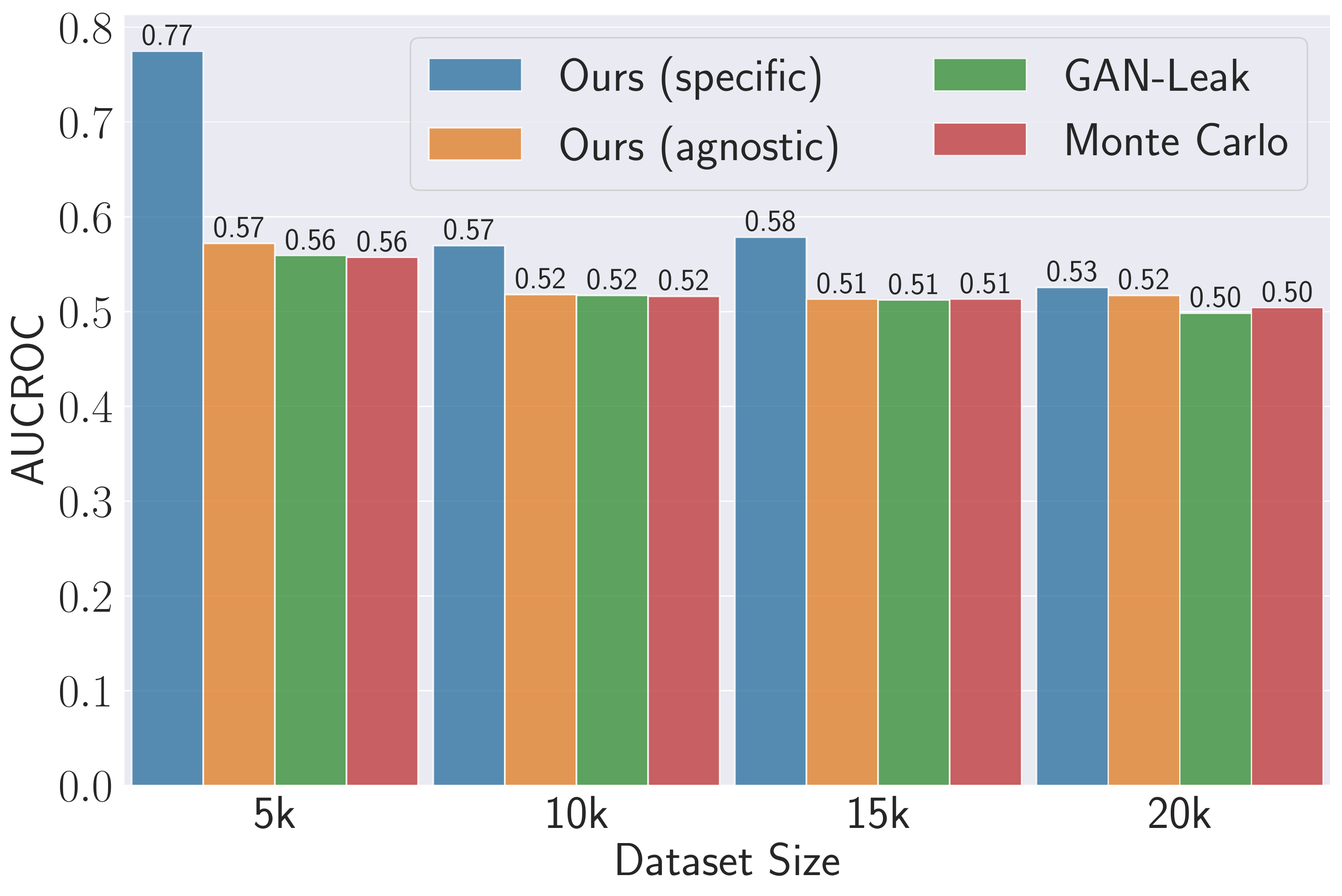}
\label{fig:blackbox_cifar_comparison}
}
\caption{The \textbf{black-box} attack AUCROC on \ref{fig:blackbox_celeba_comparison} CelebA and \ref{fig:blackbox_cifar_comparison} CIFAR-10, respectively. We adopt the gray-box attack (with truncation) on the shadow model for our model-specific attack.}
\label{fig:blackbox_comparison}
\end{figure*}

\begin{table}[!t]
\vspace{6pt}
\centering
\definecolor{Gray}{gray}{0.9}
\newcommand{\cc}{\cellcolor{Gray}}
\resizebox{\columnwidth}{!}{%
\aboverulesep=0ex
\belowrulesep=0ex
\begin{tabular}{c|cccc|cccc}
\toprule
& \multicolumn{4}{c|}{5k} & \multicolumn{4}{c}{20k} \\
\midrule
 & Median & Sum & Min & Max & Median & Sum & Min & Max \\ 
\midrule
w/o & 0.62 &0.56& 0.56& 0.49 &0.53&0.50&0.49 &0.50 \\ 
\rowcolor{Gray}
w &  \textbf{0.91}& 0.65&0.56 &0.52& \textbf{0.65}& 0.54 &0.49 &0.50\\
\bottomrule
\end{tabular}%
}
\caption{The \textbf{gray-box} attacks AUCROC on CelebA under wrong guessing of (the scheduler for) $\alpha_t$ with (``w'') or without (``wo'') applying the truncation technique. The truncation step is set to be the default value $T_{trun}=0.75T$. We highlight the best performance in each configuration in bold.}
\label{tab:greybox_rescheduler}
\end{table}

\begin{table}[!t]
\vspace{6pt}
\centering
\definecolor{Gray}{gray}{0.9}
\newcommand{\cc}{\cellcolor{Gray}}
\resizebox{\columnwidth}{!}{%
\aboverulesep=0ex
\belowrulesep=0ex
\begin{tabular}{c|cccc|cccc}
\toprule
& \multicolumn{4}{c|}{75\%} & \multicolumn{4}{c}{50\%} \\
\midrule
 & Median & Sum & Min & Max & Median & Sum & Min & Max \\ 
\midrule
w/o & \textbf{1.00} & 0.69 & 0.61&0.51 & \textbf{1.00} &0.69 &0.57 &0.50  \\ 
\rowcolor{Gray}
w &  \textbf{1.00}
&0.94 &0.61 & 0.50 & \textbf{1.00} &0.94 &0.57 &0.50\\
\bottomrule
\end{tabular}%
}
\caption{The \textbf{gray-box} attacks AUCROC on CelebA with 5k training samples when the model owner suppresses the intermediate output (with $75\%$ or $50\%$ suppression ratio). We evaluate both with (``w'') and without (``wo'') the truncation technique, where the truncation step is always set to $T_{trun}=0.75T$. We highlight the best performance in each configuration in bold.}
\label{tab:greybox_respace}
\end{table}

\subsection{Evaluation on Black-box attack}
\label{subsec:eval_blackbox}

\myparagraph{Model-Specific Attack and Cross-model Generalization}
The model owner may decide to hide intermediate results when deploying an API, limiting the attacker's access to only the final synthetic output and limiting control over the generation process. In extreme cases, the attacker may only have access to the final synthetic output without any intermediate results. However, such APIs may still provide clues about the underlying model used~\cite{ramesh2022hierarchical,saharia2022photorealistic}. In such cases, training a shadow model as a proxy of the target and conducting the attack on the shadow model would be a good strategy.

With a proxy model in hand, the attacker can apply either white-box or gray-box attack techniques discussed previously. We present results using the default settings for both white-box and gray-box attacks with truncation in \figureautorefname~\ref{fig:bb_size}. We observe that the attack performance decreases as the dataset size increases, but reasonable levels of AUCROC values above 0.6 are still generally obtained. %for datasets smaller than 10k. 
The performance of the gray-box and white-box attacks is generally comparable. By default, we use the gray-box attack with truncation for further evaluation due to its overall stability.

We take a step further into the investigation of the cross-architecture generalization of our model-specific black-box attack. Specifically, we study the scenario where the shadow model has a different architecture and may adopt a different setup of the key hyperparameters than the target model. As seen in Table~\ref{tab:blackbox_cross_architecture}, there is a slight decrease in attack AUCROC (0.77) compared to when the shadow model and target model have the same architecture (AUCROC 0.82). Furthermore, changing the key hyperparameter (the denoising step in our case) also results in a slight decrease in AUCROC from 0.77 to 0.73, but the change is not substantial. This aligns with previous research findings,  as changing the architecture can cause the shadow model to be less similar to the target model.

However, the difference in architecture may have less impact on attacks against generative models compared to classification models. In the black-box scenario, membership information in generative models is completely contained in the generated distribution, which can still be captured by a shadow model with a different architecture. In contrast, for classification models, membership information is mainly represented by their specific responses to each query, which can vary greatly between models with different architectures. Therefore, attacks based on shadow models remain relatively effective in cross-architecture scenarios for generative models, unlike MIAs against classification models that tend to become less effective~\cite{hayes2019logan,shokri2017membership}.

\begin{table}[!ht]
\centering
\aboverulesep=0ex
\belowrulesep=0ex
\resizebox{\columnwidth}{!}{%
\begin{tabular}{c|cccc}
\toprule
Diffusion Steps & AUCROC   & Accuracy & F1-Score & TPR@1$\%$FPR \\
\midrule
2000            & 0.73		& 0.68	& 0.68 & 5.99\% \\ 
4000            & 0.77		&0.69	&0.69  & 7.34\% \\ 
6000            & 0.76		&0.68	&0.68  & 6.77\% \\ \bottomrule
\end{tabular}%
}
\caption{The \textbf{black-box} attack performance when attacking a \textit{guided diffusion model} with an \textit{improved diffusion model} as the shadow model. Different settings of the diffusion steps are considered when training the shadow model. The experiments were carried out on the CelebA dataset with 5k training samples.}
\label{tab:blackbox_cross_architecture}
\end{table}

\begin{comment}
\begin{figure*} [!h]
	\centering
	\subfigure[Guided Diffusion]{
 \includegraphics[width=.22\textwidth]{figures/guided_diffusion_img_samples.pdf}
		\label{fig:guided_diffusion}}
	\subfigure[Improved Diffusion]{
		\includegraphics[width=.22\textwidth]{figures/improved_diffusion_img_samples.pdf}
		\label{fig:improved_img_samples}}
	\subfigure[StyleGAN]{
		\includegraphics[width=.22\textwidth]{figures/style_gan_img_samples.pdf}
		\label{fig:stylegan_img_samples}}
	\subfigure[PGGAN]{
		\includegraphics[width=.22\textwidth]{figures/pggan_img_samples.pdf}
		\label{fig:pggan_img_samples}}
	\caption{The synthetic images sampled from Guided Diffusion, Improved Diffusion, StyleGAN and PGGAN trained on Celeba with 5k samples, respectively.}
\label{fig:imgs_samples_from_different_generative_models}
\end{figure*}
\end{comment}

\myparagraph{Model-based vs. Model-agnostic}
For the least informed attack scenario, attackers would have to rely on a model-agnostic method, i.e., they cannot use any extra knowledge of the target model except blindly collecting generated samples from it. In this case, we build upon existing methods that calculate the distance between the query sample and generated samples (closer distance indicates higher membership probability). We improve upon these methods by enhancing the distance metrics, i.e., we use a pre-trained ImageNet classifier as a feature extractor and compute the cosine distance between features as the metrics. Our modification leverages the rich semantic information from the pre-trained feature extractor to improve the discriminative power of the resulting membership score. The comparisons to existing methods are shown in Figures~\ref{fig:blackbox_celeba_comparison} and \ref{fig:blackbox_cifar_comparison}. As shown, our model-agnostic attack performs slightly better than existing methods, while our model-specific attack greatly improves by leveraging slightly more information that is always freely available even in a black-box setting.

%As mentioned in xx, a shadow model is necessary for conducting black-box attacks. In our experiments, we evaluated two different MIA attack strategies: feature-based KNN and shadow model-based \textit{DiffMIA}. With the feature-based KNN approach, we use a model trained on public data to transform the original data into a dense embedding. We then calculate the cosine distance between the embeddings of test samples and released samples, which serves as a metric for determining the likelihood that a sample was used to train the generative models.

% \begin{table}[]
% \resizebox{\columnwidth}{!}{%
% \begin{tabular}{c|ccccc|c|c}
% \hline
%               Size of    & \multicolumn{5}{c|}{Truncated MSE Loss}     & GAN-Leak & Monte-Carlo \\ 
% Train Set & min    & max    & Median & avarage & sum    &          &             \\ \hline
% 4096              & 0.5082 & 0.5031 & 0.7815 & 0.6471  & 0.6471 & 0.662    & 0.67        \\ \hline
% 10000             & 0.5013 & 0.5031 & 0.6393 & 0.5783  & 0.5783 & 0.5421   & 0.5415      \\ \hline
% 15000             & 0.4969 & 0.4966 & 0.5969 & 0.5562  & 0.5563 & 0.511    & 0.511       \\ \hline
% 20000             & 0.491  & 0.4872 & 0.5881 & 0.5467  & 0.5467 & 0.493    & 0.497       \\ \hline
% \end{tabular}%
% }
% \caption{The attack AUC for Truncated MSE Loss and the SOTA GAN-Leak and Monte-Carlo approaches with different size of training set. }
% \label{tab:vlb_black_box_setting}
% \end{table}

\subsection{Exploring Larger Dataset Sizes}
\label{subsec:large_dataset}
 \begin{figure*}[!t]
     \centering
     \subfigure[white-box (CelebA)]{ 
 \includegraphics[width=0.23\textwidth]{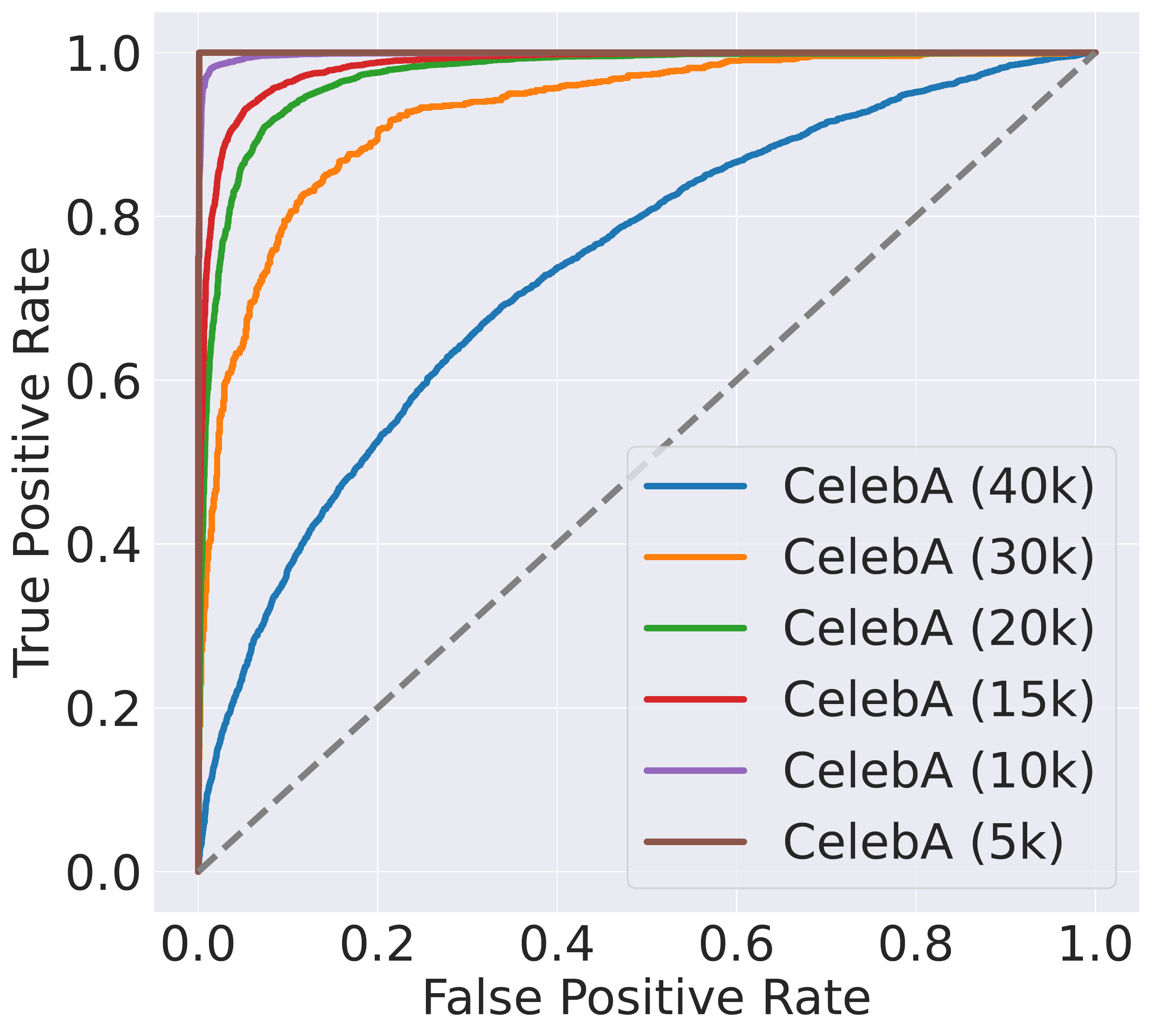}\label{fig:whitebox_roc_celeba}
     }
          \subfigure[gray-box (CelebA)]{ 
 \includegraphics[width=0.23\textwidth]{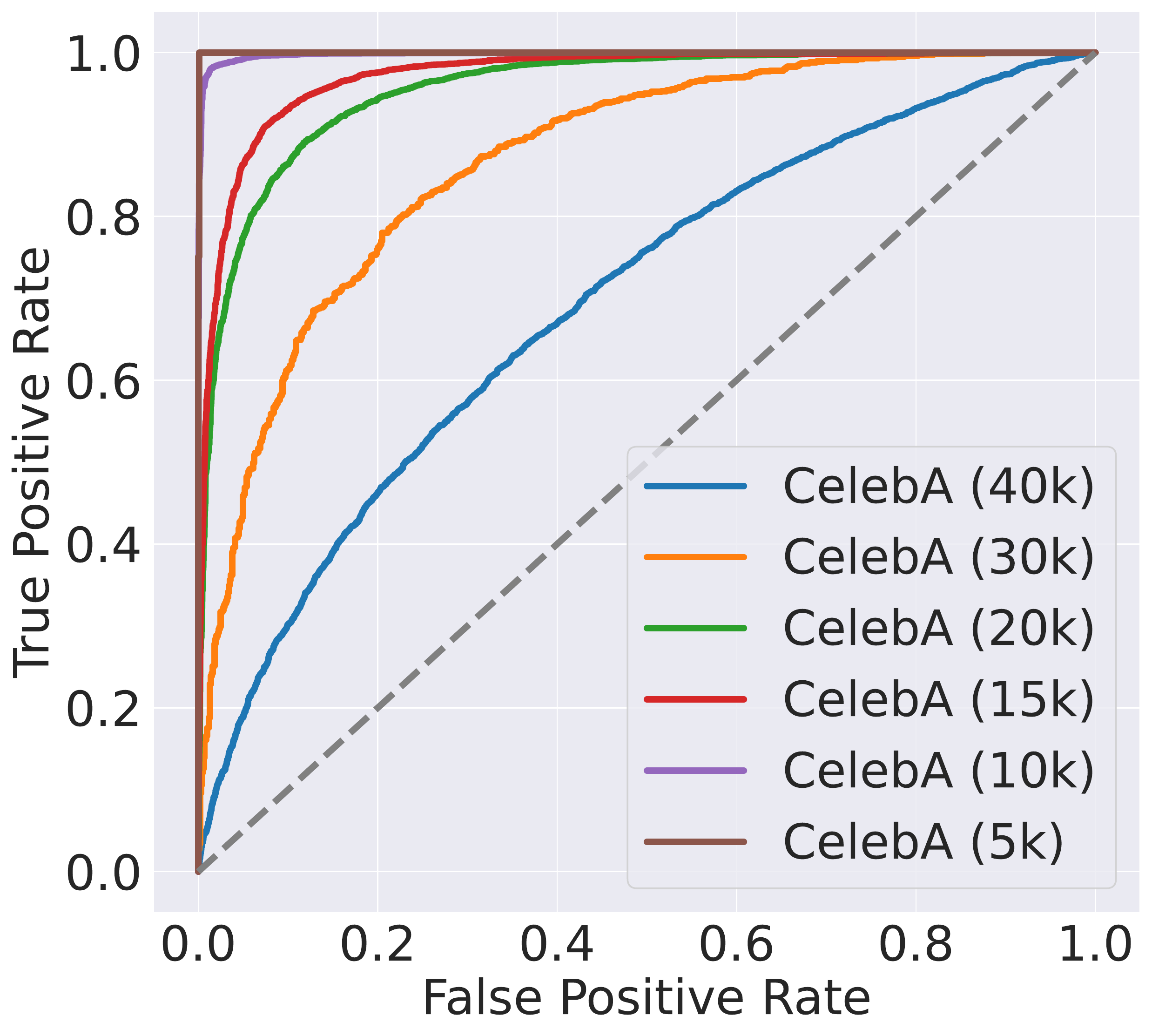}\label{fig:graybox_roc_celeba}
     }
          \subfigure[white-box (CIFAR-10)]{ 
 \includegraphics[width=0.23\textwidth]{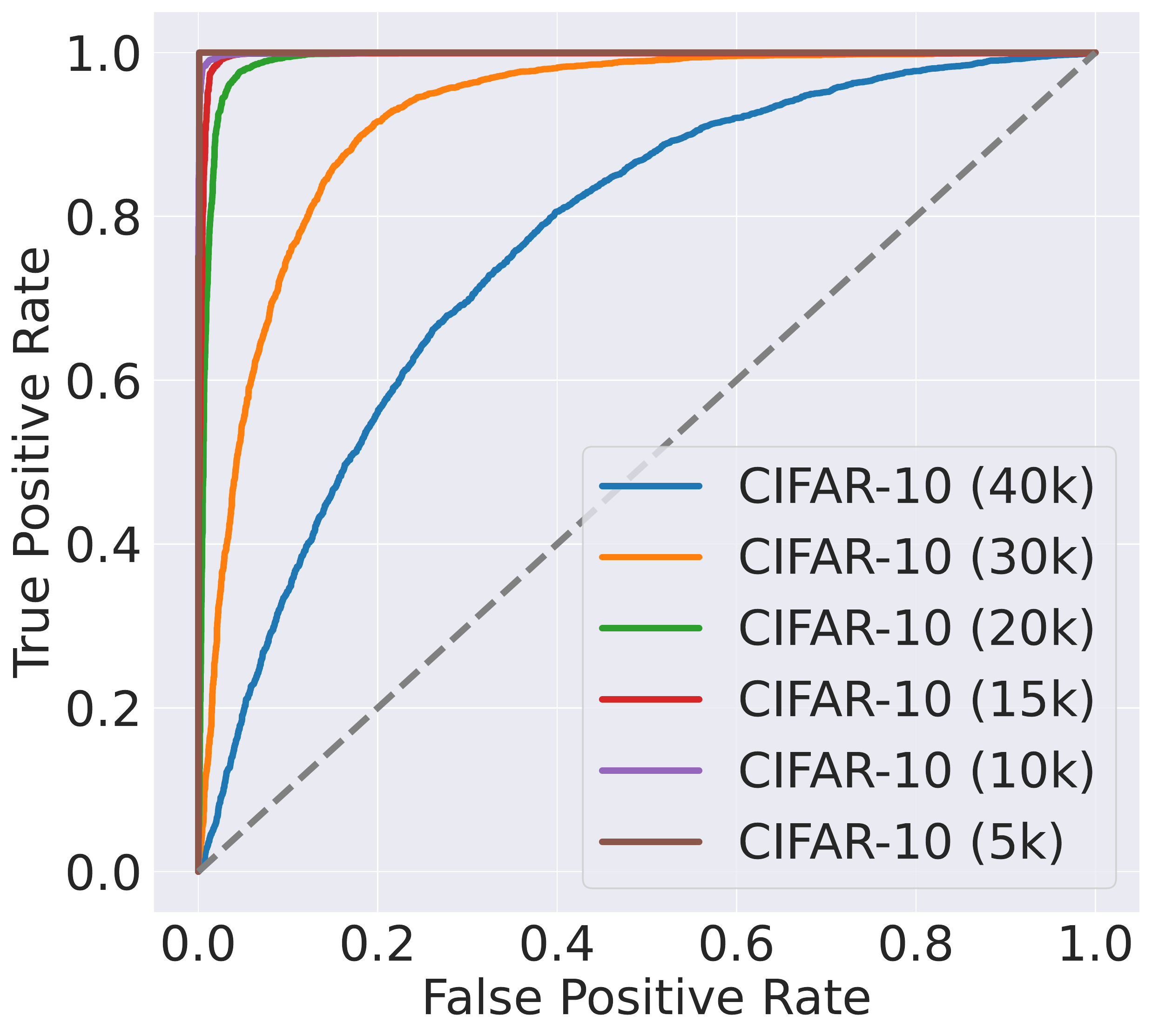}\label{fig:whitebox_roc_cifar}
     }
    \subfigure[gray-box (CIFAR-10)]{ 
 \includegraphics[width=0.23\textwidth]{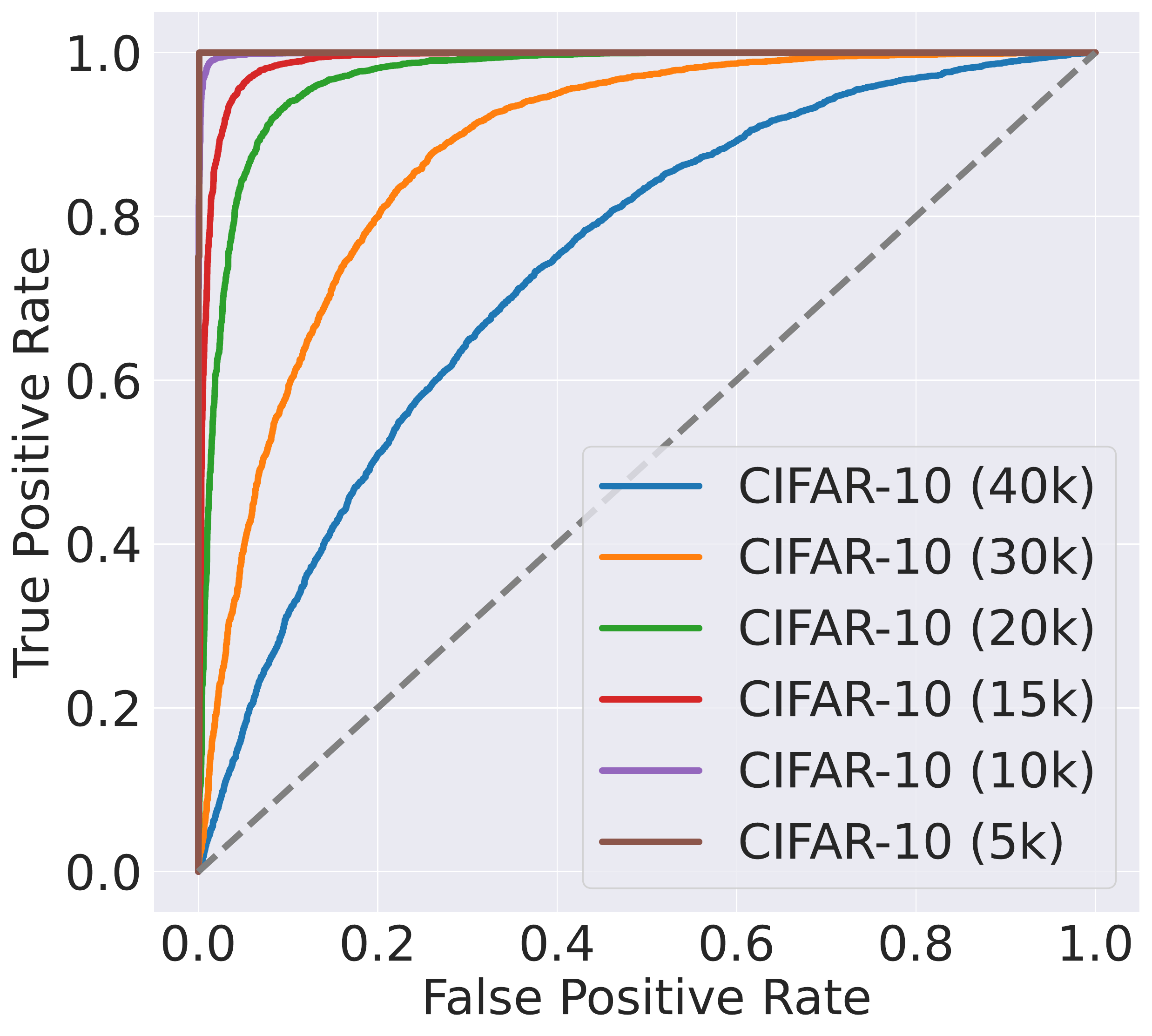}\label{fig:graybox_roc_cifar}
     }
     \caption{ROC curves of our \textbf{white-box} and \textbf{gray-box} (with truncation techniques) attacks on CelebA and CIFAR-10. }
 \end{figure*}
 
% \begin{figure}[!t]
% \centering
% \includegraphics[width=0.9\columnwidth]{figures/graybox_truncation_different_dataset_size_roc_celeba.pdf}
% \caption{ROC curves of \textbf{gray-box} attacks (with truncation techniques) on CelebA across different dataset sizes. }
% \label{fig:greybox_rocs}
% \end{figure}

% While previous studies have typically reported effective attacks on datasets containing no more than 20k training data points, our attack performance has not yet reached its saturation point in such scenarios, maintaining an AUCROC of over 0.95 (for both gray-box and white-box cases). To explore the limits of our approach, we conducted further experiments using guided diffusion models trained with larger volumes of data.Notably, a training set with $>$20k samples is sufficiently large to effectively train a diffusion models. As shown in Table~\ref{tab:larger_dataset_graybox_cifar10}, our approach remains capable of extracting membership information, with an AUC exceeding 0.7. When the false-positive rate is reduced to 0.1\%, the true positive rate remains at 0.36\%. This indicates that even under these conditions, membership information can still be compromised. Importantly, this leakage occurs (0.36\% TPR) even when adversaries operate an MIA with a low false-positive rate (0.1\%).
While previous studies  mainly reported effective attacks on datasets with no more than 20k training data samples, our attack performance has not yet reached its saturation point in such scenarios, maintaining an AUCROC of over 0.95 (for both gray-box and white-box cases). To push the boundaries of our approach, we further experimented with guided diffusion models trained with a larger size of training data (up to the maximum effective size for the adopted benchmarking datasets that still leaves a sufficient amount of data for MIA evaluation). As presented in Table~\ref{tab:larger_dataset_graybox_cifar10}, our approach is still capable of extracting membership information, achieving an AUCROC  that generally exceeds $0.7$. Furthermore, the attack maintains a high TPR under specific low FPR thresholds (0.1\% and 1\%). Specifically, the TPR is always substantially higher than the FPR at least three times greater, indicating that the attack could successfully identify significantly more member samples than it misclassified non-members as members (refer to \figureautorefname~\ref{fig:whitebox_roc_celeba} for the complete ROC curves), which points to a strong attack. Notably, an MIA is deemed successful ``if it can reliably violate the privacy of even just
a few users in a sensitive dataset''~\cite{carlini2022membership}.

\begin{table}[!ht]
\centering
\aboverulesep=0ex
\belowrulesep=0ex
\resizebox{\columnwidth}{!}{%
\begin{tabular}{c|c|ccc}
\toprule
Dataset & Dataset Size & AUCROC   & TPR@0.1$\%$FPR & TPR@1$\%$FPR \\
\midrule
\multirow{3}{*}{CIFAR-10}  
& 20k  & 0.97 & 8.34\% & 41.57\%  \\ 
& 30k  & 0.87 & 1.52\% & 10.06\% \\ 
& 40k  & 0.73 & 0.36\% &  4.48\% \\ 
\midrule
\multirow{3}{*}{CelebA}  
& 20k & 0.98 & 15.22\% & 51.06\% \\
& 30k & 0.91 & 8.19\%  & 34.14\% \\
& 40k & 0.69 & 1.18\% & 6.06\%   \\
\bottomrule
\end{tabular}%
}
\caption{The \textbf{gray-box} attack performance against guided diffusion models with varying dataset size.}
\label{tab:larger_dataset_graybox_cifar10}
\end{table}

\subsection{MIA Comparison between Diffusion Models and GANs}
\label{subsec:exp_different_gm}
Our results presented in previous sections suggest that diffusion models may be generally more vulnerable to MIAs in comparison to other popular generative models such as GANs under various attack scenarios.
In this section, we closely examine their behavior under MIAs in a similar setting. We train and evaluate both diffusion and GAN models on the same sample set using the same attack method (i.e., the model-agnostic black-box attack from \cite{chen2020gan}). We consider the widely used models that represent the state-of-the-art or previous state-of-the-art at their release time. These include two GAN models, namely the StyleGAN~\cite{karras2019style} and progressive GAN (PGGAN)~\cite{karrasprogressive}, and the ``Guided'' and ``Improved'' diffusion models. The results in \tableautorefname~\ref{tab:comparison_gan_diffusion} show that diffusion models have higher attack AUCROC compared to GAN models, with PGGAN having 0.56 and Improved Diffusion having 0.62, while the difference in the TPR@1\%FPR becomes much more significant (one magnitude more TPR when attacking diffusion models). These also support that diffusion models tend to be more vulnerable to MIA attacks than GAN models, even when considering only model-agnostic attacks, not to mention the exceptionally high privacy risk when our dedicated approaches are applied.

\begin{table}[!ht]
\centering
\begin{adjustbox}{max width=\columnwidth}
\resizebox{\columnwidth}{!} {
\aboverulesep=0ex
\belowrulesep=0ex
\begin{tabular}{c|cccc}
\toprule
 & Guided & Improved & StyleGAN & PGGAN \\
\midrule
FID        & 22.46            & 24.78      & 25.89    & 55.28  \\
\midrule
AUCROC & 0.59           & 0.62    & 0.51    & 0.57            \\
TPR@1$\%$FPR & 5.14\% & 6.11\% & 1.02\% & 4.04\% \\
\bottomrule
% Guided     & 22.46       & 0.59     & 10   \\
% Improved   & 24.78       & 0.62     & 14   \\ 
% StyleGAN   & 25.89       & 0.51     & 1   \\ 
% PGGAN      & 55.28       & 0.57     & 4   \\ 
% \bottomrule
\end{tabular}%
}
\end{adjustbox}
\caption{The generation quality (FID) and attack performance (AUCROC, TPR@1\%FPR) of various generative models trained on the CelebA dataset with 5k samples. The model-agnostic attack from \cite{chen2020gan} was evaluated. The FID was calculated based on 40k generated images from each generative model.}
\label{tab:comparison_gan_diffusion}
\end{table}

\begin{comment}
\begin{table}[!ht]
\centering
%\begin{adjustbox}{max width=\columnwidth}
\resizebox{\columnwidth}{!} {
\aboverulesep=0ex
\belowrulesep=0ex
\begin{tabular}{c|cccc}
\toprule
 & Guided & Improved & StyleGAN & PGGAN \\
\midrule
FID        & 22.46            & 24.78      & 25.89    & 55.28  \\
\midrule
GAN-Leak   & 0.59           & 0.62    & 0.51    & 0.57            \\ 
MC & 0.59             & 0.61       & 0.51    & 0.56   
        \\ \bottomrule
\end{tabular}%
}
%\end{adjustbox}
\caption{The attack AUCROC and the FID of various generative models trained on the CelebA dataset using 5k training samples. The FID was calculated based on 40k generated images from each generative model.}
\label{tab:comparison_gan_diffusion}
\end{table}
\end{comment}

\section{DISCUSSION}
\label{sec:discussion}
In this section, we highlight several key insights and their practical implications, as well as discuss possible concerns regarding our attack formulation.

\subsection{Privacy Risks of Diffusion Models}
\label{subsec:diss_comparison_gm}
Our results in \sectionautorefname~\ref{sec:experiments} (particularly in \sectionautorefname~\ref{subsec:exp_different_gm}) show that MIAs have a notably higher attack success rate when targeting diffusion models compared to other popular generation models like GANs across various attack scenarios. This possibly can mainly be attributed to the objective used by diffusion models. The objective, which is to maximize the log-likelihood lower bound on all training samples, can result in a loss landscape that locally minimizes the loss around each training sample, potentially leading to a spike in the  distribution if not properly regularized. 
 This inevitably leaves clues for attackers to successfully conduct their attacks. In contrast, the adversarial objective used in GANs indirectly guides the generator to produce samples that resemble the training data, while also preventing exact memorization through adversarial updates. These indicate that diffusion models may intrinsically pose a higher privacy risk and should be used with caution in real-world applications, especially considering their widespread use as a standard media generation tool. 
 
Additionally, it is relatively easy to reduce privacy risk for other generative models by only releasing the functional part (e.g. the generator) and keeping the unnecessary part (e.g. the encoder in VAEs or the discriminator in GANs) private~\cite{chen2020gan}. However, this is normally not the case for diffusion models, since the unnecessary part of diffusion models (the forward process) is fixed, unlearned, identical or highly similar across models and settings, making it easy for the attacker to guess and mimic the real process. As a result, an attack generally requires less effort to associate each query sample with its latent variable and estimate the likelihood needed for the attack. This can be seen in the qualitative results shown in \autoref{fig:intermediate_generation_process_guided_diffusion}: the reconstruction is more accurate when compared to previous cases that required gradient-based optimization~\cite{chen2020gan}. This characteristic of diffusion models thus poses higher potential risks in deployment scenarios.

\subsection{Conditional Generation}
\label{subsec:diss_conditional}
Our approach can be seamlessly extended to conditional generation models, such as the text-to-image stable diffusion models. 
Specifically, we consider each query sample as a combination of an image and its accompanying text description. We extract multiple images generated at various diffusion steps from the target diffusion models for the query text, and derive the membership score via Equation~\ref{eq:grey_box_attack}. However, the loss is computed only on the image component $\vx^i_{\mathrm{img}}$, and the text $\vx^i_{\mathrm{text}}$ serves as an additional input to the model.
We investigate two publicly released pre-trained models, stable-diffusion-v1.4\footnote{\href{https://huggingface.co/CompVis/stable-diffusion-v1-4}{https://huggingface.co/CompVis/stable-diffusion-v1-4}}  and stable-diffusion-v1.5\footnote{\href{https://huggingface.co/runwayml/stable-diffusion-v1-5}{https://huggingface.co/runwayml/stable-diffusion-v1-5}}, both from the official Huggingface repository. We use the inference implementation without any modifications to collect the generated images at different steps, to simulate the gray-box setting. For evaluation, we use COCO-2017~\cite{lin2014microsoft} as the non-member set, which exhibits a distribution similar to the member set (i.e., Laion2B-improved-aesthetics~\cite{schuhmannlaion}) used by the pre-trained stable diffusion models. All images are resized to a resolution of 512$\times$512 for evaluation. As Table~\ref{tab:mia_against_pretrained_stable_diffusion} demonstrates, our approach effectively extracts membership information from these real-world diffusion models, trained on large-scale dataset, specifically, with 2.3 billion samples. Remarkably, our method maintains a significant level of TPR (exceeding 24\%) even at a low FPR of 1\%. This demonstrates its potential in accurately tracing the usage of specific samples during the training of a diffusion model, while concurrently highlighting the privacy risks associated with potential training data leakage when deploying or sharing such models that operate on sensitive data.

% To highlight the vulnerability of diffusion-based generative models against MIA, we conduct the evaluation on the text-to-image models, e.g. stable diffusion. Our approach can easily be extended to it. We extract multiple images generated from different steps for a given text, then apply our approach to obtain the membership scores. In this evaluation, we investigate two released pre-trained models, stable-diffusion-v1.4~\footnote{\href{https://huggingface.co/CompVis/stable-diffusion-v1-4}{https://huggingface.co/CompVis/stable-diffusion-v1-4}}  and stable-diffusion-v1.5~\footnote{\href{https://huggingface.co/runwayml/stable-diffusion-v1-5}{https://huggingface.co/runwayml/stable-diffusion-v1-5}}, both from Huggingface official version and apply the inference implementation without any change to collect the generated images for different steps. In the evaluation, we take the \textbf{Laion2B-improved-aesthetics} as the member set, while \textbf{COCO-2017} as the non-member set.Table~\ref{tab:mia_against_pretrained_stable_diffusion} shows that our approach is effective to extract the membership information from the released large diffusion model trained on larger training data. Even in the low FPR, e.g., 1$\%$, our approach still achieves a 24.21$\%$ TPR, which highlights the seriousness of training data leakage in diffusion-based generative models. 
% \usepackage{graphicx}
\begin{table}[!ht]
\centering
\aboverulesep=0ex
\belowrulesep=0ex
\resizebox{\columnwidth}{!}{%
\begin{tabular}{c|ccc}
\toprule
Models & AUCROC   & F1-Score & TPR@1$\%$FPR \\
\midrule
stable-diffusion-v1.4           & 0.73	& 0.71 & 24.21 $\%$ \\ 
stable-diffusion-v1.5            & 0.74	& 0.71 & 25.66 $\%$ \\ 
\bottomrule
\end{tabular}%
}
\caption{Our attack performance on the pre-trained stable diffusion models from Huggingface.}
\label{tab:mia_against_pretrained_stable_diffusion}
\end{table}

% While this work focuses on image generation models, exploring possible extensions to popular conditional generation tools, such as text-to-image~\cite{saharia2022photorealistic,ramesh2022hierarchical,rombach2022high} and image-to-image~\cite{dhariwal2021diffusion} models, is an interesting direction. In the white-box case where the model internals are accessible, our attack can be easily applied to the image generation part if only the image embedding determines the generation process. However, if the generation part requires both image embedding and additional information, mapping the image embedding to the space of the conditioning information, such as text embedding, may be necessary. This can be achieved using pre-trained models, such as image captioning to generate descriptive text from query images or the CLIP model~\cite{radford2021learning} to match the image embedding with the text embedding in a semantic preserving way. The same situation applies to the gray-box setting, which typically requires such extra mappings. Recent works have shown promising results using pre-trained models for deepfake detection in diffusion models~\cite{sha2022fake}, and we expect positive outcomes in the task of membership inference. However, more in-depth investigation is needed and is left for future work as it requires dedicated effort to reverse engineer the appropriate mapping given the target conditional generation method's implementation information and is orthogonal to our contributions in this work.

% \vspace{-7pt}
\subsection{Potential Defenses}
\label{subsec:diss_defense}

As presented in \sectionautorefname~\ref{subsec:eval_blackbox}, limiting the information available to attackers is generally effective in protecting against such attacks. A slight decrease in attack performance can occur when the model developer hides important parameters, causing the attacker to make incorrect guesses, while a larger degradation happens when the model owner further prevents controllable generation (comparing gray-box to black-box model-specific attacks) and even obscures the sources of synthetic samples (comparing model-specific to model-agnostic attacks). However, these measures can come at the cost of a degraded user experience and may not be a sustainable solution.

Providing rigorous privacy guarantees is another option for the defense. Differential privacy (DP)~\cite{dwork2014algorithmic} is a widely used technique that ensures protection against privacy attacks. To prevent privacy leakage from machine learning models, DP incorporates adding random noise to the gradients during training to reduce the impact of each individual sample on the model parameter and thus hide the presence of the data in the training set~\cite{abadi2016deep}. However, DP training inevitably hampers the model utility and significantly increases the computational cost during training. While notable recent progress has been achieved in developing DP generative models, these advancements are largely limited to simple datasets like MNIST and Fashion-MNIST, and do not offer a practical solution for the complex datasets considered in this work (for example, see generation results in  \cite{dockhorn2022differentially}). We believe a more in-depth investigation into developing efficient and effective defense mechanisms for MIA on diffusion models is required but leave it as future work as it is orthogonal to our contributions in this work.

\section{CONCLUSION}
In this work, we present the first systematic analysis of membership inference attacks against diffusion models. Our study presents, for the first time, the key attack vectors that are particularly relevant  for real-world deployment scenarios of diffusion models. Moreover, we propose our novel attack approaches tailored to each attack scenario. Our methods exploit readily available information while delivering promising performance across a broad range of settings, thereby demonstrating high potential for application scenarios that necessitate accurate auditing of data usage when developing and deploying diffusion models. %Our results highlight the high potential privacy risk associated with diffusion models with which we aim to motivate further efforts into related topics.
Our findings, coupled with our insights, highlight the high potential privacy risks associated with diffusion models, an area we believe warrants further exploration. To facilitate future research in this field, the source code implementation will be made openly available upon publication.


% \section*{Acknowledgment}



%\bibliographystyle{IEEEtranS}
\begin{comment}

\end{comment}
% \bibliographystyle{IEEEtranS}
\bibliographystyle{abbrv}
\bibliography{ndss2024}

\clearpage
% \appendix
  \appendices
\section{Experiment Configuration}
\subsection{Setup}
We present the additional details of our experimental setup in this section. Table~\ref{tab:model_hyperparameters} summarizes the key hyperparameters that we adopted in training the guided diffusion and improved diffusion models.  For the stable diffusion model, we use the official released models with the same hyper-parameters from the Huggingface website\footnote{\url{{https://huggingface.co/CompVis}}}. For StyleGAN\footnote{\url{https://github.com/NVlabs/stylegan}} and PGGAN\footnote{\url{https://github.com/tkarras/progressive_growing_of_gans}}, we use the official open-sourced implementation with the default hyperparameters for training. All experiments were conducted on a single NVIDIA A100 GPU. %We use the officially released implementation to train the models with the default parameters setting. 

\begin{table}[!h]
\centering
%\resizebox{\columnwidth}{!}{%
% \resizebox{\columnwidth}{!}{%
\begin{tabular}{ccccc}
\hline
Hyperparameters & \begin{tabular}[c]{@{}c@{}}Guided\\  Diffusion\end{tabular} & \begin{tabular}[c]{@{}c@{}}Improved \\ Diffusion \end{tabular} &  StyleGAN & PGGAN \\ \hline
channels        & 128    & 128     & 512  & 128    \\ 
residual block  & 3      & 3       & 3  & -    \\ 
learn sigma     & True   & True    & -  & -     \\ 
noise scheduler & linear & linear  & -  & -     \\ 
batch size      & 256    & 256     & 64  & 64     \\ 
learning rate   & 1e-3   & 1e-3    & 1e-3  & 1e-3     \\ 
diffusion steps & 4000   & 2000;4000;6000 & - & - \\ 
dropout         & 0.3    & 0.3          & 0.3  & 0.3 \\ \hline
\end{tabular}%
% }
\caption{Summary of the Training Hyperparameters.}
\label{tab:model_hyperparameters}
\end{table}

\subsection{Dataset}

\myparagraph{Laion2B-improved-aesthetics\footnote{\href{https://huggingface.co/datasets/laion/laion2B-en-aesthetic}{https://huggingface.co/datasets/laion/laion2B-en-aesthetic}}}
The Laion2B-improved-aesthetics is a curated subset of Laion2B that focuses on images with high-resolution quality and improved aesthetics. It consists of color images with resolutions of 512$\times$512 or higher. Each image has an estimated aesthetics score of $>$5.0 and an estimated watermark probability of $<$0.5. Furthermore, the text caption for each image in Laion2B-improved-aesthetics is in English. The dataset includes more than 2.3 million image-caption pairs.
% The Laion2B-improved-aesthetics is a curated subset of Laion2B focusing on images with high-resolution quality and improved aesthetics. It consists of images with resolutions $>=$ 512x512. Each image has an estimated aesthetics score of $> 5.0$ and an estimated watermark probability of $<$ 0.5. Besides, the caption of Laion2B-improved-aesthetics is English. The dataset includes more than 1 million image and caption pairs.

\myparagraph{COCO-2017\footnote{\href{https://cocodataset.org/}{https://cocodataset.org/}}}
COCO is a large-scale object detection, segmentation, and captioning dataset. It contains more than 200k images and 80 object categories. Each image is associated with one annotated English text caption. We randomly sample the images from datasets and resize it from the original resolution to 512$\times$512 using the open-source scripts\footnote{\href{https://github.com/rom1504/img2dataset}{https://github.com/rom1504/img2dataset}}. %With the descriptive caption for each image, the dataset enables researchers to explore and advance various computer vision tasks, e.g., object segmentations, Superpixel stuff segmentation, etc. 

\section{Additional Results}
\subsection{Generation Quality}
\begin{figure*} [!t]
	\centering
	\subfigure[Guided Diffusion]{
 \includegraphics[width=.22\textwidth]{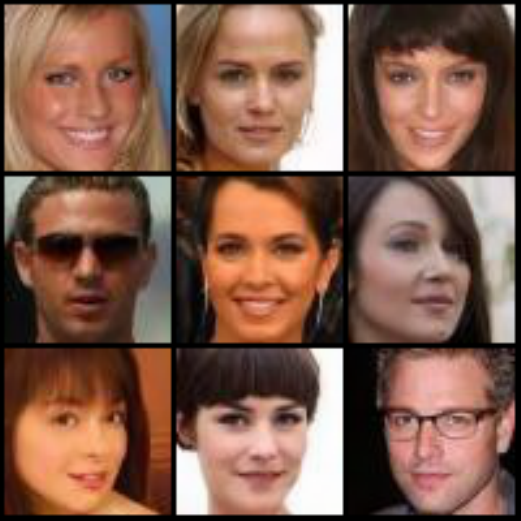}
		\label{fig:guided_diffusion}}
	\subfigure[Improved Diffusion]{
		\includegraphics[width=.22\textwidth]{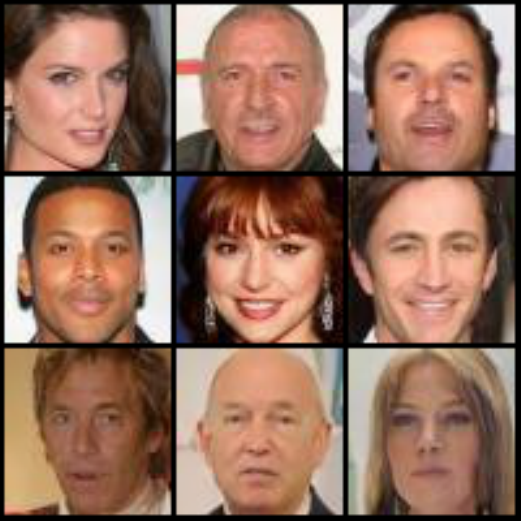}
		\label{fig:improved_img_samples}}
	\subfigure[StyleGAN]{
		\includegraphics[width=.22\textwidth]{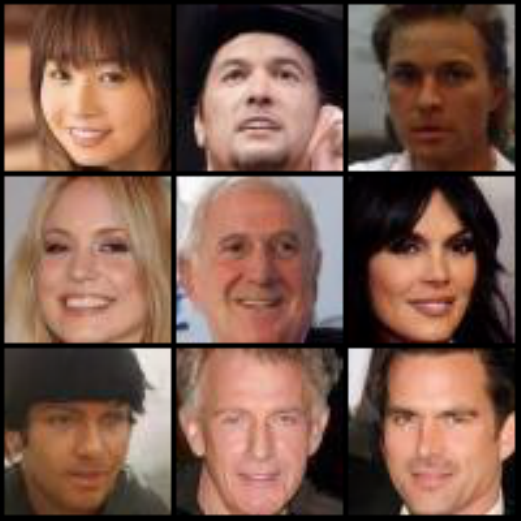}
		\label{fig:stylegan_img_samples}}
	\subfigure[PGGAN]{
		\includegraphics[width=.22\textwidth]{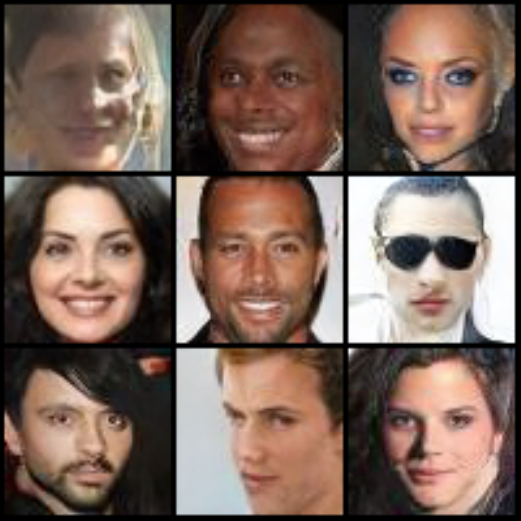}
		\label{fig:pggan_img_samples}}
	\caption{The synthetic images sampled from Guided Diffusion, Improved Diffusion, StyleGAN and PGGAN trained on Celeba with 5k samples, respectively.}
\label{fig:imgs_samples_from_different_generative_models}
\end{figure*}

We display the generated samples from different generative models in Figure~\ref{fig:imgs_samples_from_different_generative_models}. As can be observed, all models demonstrate a reasonable level of generation quality (practical utility), and none of the models exhibit significant visual differences in their generation quality. This consistency controls the factor of generation quality in their vulnerability to MIAs. The associated quantitative measurements (e.g., FID) are presented in Table~\ref{tab:comparison_gan_diffusion}.

\subsection{White-box Setting}

We present the investigation of various statistic functions $f$ on the loss trajectory on CIFAR-10 dataset in  \tableautorefname~\ref{tab:cifar10_wb_function}. The results confirm the consistency with findings from experiments on the CelebA dataset.
\begin{table}[!ht]
\centering
%\begin{adjustbox}{max width=\columnwidth}
\resizebox{\columnwidth}{!}{%
\definecolor{Gray}{gray}{0.9}
\newcommand{\cc}{\cellcolor{Gray}}
\aboverulesep=0ex
\belowrulesep=0ex
\centering
\begin{tabular}{c|c|cccc}
\toprule
Size & Truncation & Min & Max & Median & Sum \\
\midrule
\multirow{2}{*}{5000}  & without & 0.51  & 0.54  & 1.00   & 0.93  \\ 
& \cc with & 0.50 \cc  & 1.00 \cc & 0.97 \cc& 1.00 \cc \\
\hdashline
\multirow{2}{*}{10000}  & without & 0.49 & 0.50 & 0.93  & 0.74 \\ 
&  \cc with & 0.49 \cc & 0.99 \cc & 0.74 \cc& 0.98 \cc \\
\hdashline
\multirow{2}{*}{15000}  & without & 0.50 & 0.50 & 0.91  & 0.73 \\
& \cc with & 0.50 \cc & 0.99 \cc & 70 \cc& 0.98 \cc \\
\hdashline
\multirow{2}{*}{20000} & without & 0.49 & 0.50 & 0.85  & 0.67 \\ 
& \cc with & 0.49 \cc & 0.99 \cc & 0.64 \cc& 0.95 \cc \\
\hdashline
\multirow{2}{*}{30000} & without & 0.51 & 0.51 & 0.73  & 0.58 \\ 
& \cc with & 0.51 \cc & 0.92 \cc & 0.58 \cc& 0.84 \cc \\
\hdashline
\multirow{2}{*}{40000} & without & 0.51 & 0.51 & 0.63  & 0.54 \\ 
& \cc with & 0.51 \cc & 0.76 \cc & 0.54 \cc& 0.70 \cc \\
\bottomrule
\end{tabular}%
%\end{adjustbox}
}
\caption{The \textbf{white-box} attack AUCROC when applying different statistic function $f$ (\textit{Min}, \textit{Max}, \textit{Median}, and \textit{Sum}) to the \textit{loss trajectory} $\{\Ls_{t}\}_{t=0}^T$ with and without truncations. The experiments were conducted on the CIFAR-10 dataset with various training set sizes, as indicated in the first column. For the cases where the truncation technique is applied, we set the truncation step to be the default value with $T_{trun}=0.75T$. }
\label{tab:cifar10_wb_function}
\vspace{6pt}
\end{table}
\begin{comment}
\begin{table}[!ht]
\centering
%\begin{adjustbox}{max width=\columnwidth}
\resizebox{\columnwidth}{!}{%
\definecolor{Gray}{gray}{0.9}
\newcommand{\cc}{\cellcolor{Gray}}
\aboverulesep=0ex
\belowrulesep=0ex
\centering
\begin{tabular}{c|c|cccc}
\toprule
Size & Truncation & Min & Max & Median & Sum \\
\midrule
\multirow{2}{*}{5000}  & without & 0.51  & 0.54  & 1.00   & 0.93  \\ 
& \cc with & 0.50 \cc  & 1.00 \cc & 0.97 \cc& 1.00 \cc \\
\hdashline
\multirow{2}{*}{10000}  & without & 0.50 & 0.50 & 0.91  & 0.85 \\ 
&  \cc with & 0.54 \cc & 0.66 \cc & 0.99 \cc& 0.99 \cc \\
\hdashline
\multirow{2}{*}{15000}  & without & 0.49 & 0.50 & 0.76  & 0.71 \\
& \cc with & 0.58 \cc & 0.60 \cc & 0.99 \cc& 0.99 \cc \\
\hdashline
\multirow{2}{*}{20000} & without & 0.49 & 0.68 & 0.49  & 0.62 \\ 
& \cc with & 0.49 \cc & 0.52 \cc & 0.93 \cc& 0.84 \cc \\
\bottomrule
\end{tabular}%
%\end{adjustbox}
}
\caption{The \textbf{white-box} attack AUCROC when applying different statistic function $f$ (\textit{Min}, \textit{Max}, \textit{Median}, and \textit{Sum}) to the \textit{loss trajectory} $\{\Ls_{t}\}_{t=0}^T$ with and without truncations. The experiments were conducted on the CIFAR-10 dataset with various training set sizes, as indicated in the first column. For the cases where the truncation technique is applied, we set the truncation step to be the default value with $T_{trun}=0.75T$. }
\label{tab:cifar10_wb_function}
\vspace{6pt}
\end{table}
\end{comment}

\vspace{6pt}
We present the quantitative results in~\tableautorefname~\ref{tab:whitebox_datasize}, which is supplementary to \figureautorefname~\ref{fig:wb_size} in the main paper.

\begin{table}[!ht]
\centering
\definecolor{Gray}{gray}{0.9}
\newcommand{\cc}{\cellcolor{Gray}}
\resizebox{\columnwidth}{!}{%
\aboverulesep=0ex
\belowrulesep=0ex
\begin{tabular}{c|cccc|cccc}
\toprule
& \multicolumn{4}{c}{CelebA} & \multicolumn{4}{c}{CIFAR-10} \\
\midrule
Truncation & 5k & 10k & 15k & 20k & 5k & 10k & 15k & 20k \\ 
\midrule
w/o & 1.00 & 0.94 & 0.80 & 0.77 & 1.00 & 0.93 & 0.91 & 0.85\\ 
\rowcolor{Gray}
w & 1.00 & 1.00 & 0.99 & 0.98 &1.00 & 0.99 & 0.99 &  0.99 \\
\bottomrule
\end{tabular}%
}
\caption{The \textbf{white-box} attack AUCROC across different training set sizes. Without (``w/o'') truncation, the statistic function is selected to be ``\textit{Median}'', while it is set to be ``\textit{Max}'' with (``w'') truncation. The truncation step is set to be $T_{trun}=0.75T$.}
\label{tab:whitebox_datasize}
\end{table}

Additionally, we examine in detail the potential factors that may impact the vulnerability of target diffusion models 
to MIA, such as truncating loss trajectory, statistical functions, training set size, etc. Our extended experiments on the CelebA dataset cover various training configurations, and the results are displayed in Table~\ref{tab:wb_celeba_truncation_all}.
 
\begin{table}[!h]
\aboverulesep=0ex
\belowrulesep=0ex
\centering
\definecolor{Gray}{gray}{0.9}
\newcommand{\cc}{\cellcolor{Gray}}
\centering
\subtable[CelebA (5k)]{
%\begin{adjustbox}{max width=\columnwidth}
\resizebox{0.95\columnwidth}{!}{%
\begin{tabular}{c|cccccc}
\toprule
$f$ & \cc $T$ & $0.975T$ & $0.875T$ & $0.75T$ & $0.625T$ & $0.5T$ \\
\midrule
Median & \cc1.00 & 1.00  & 0.99 & 0.94 & 0.77 & 0.56  \\
Sum    & \cc0.99 & 0.71 & 1.00 & 1.00 & 1.00 & 0.97  \\
Min    & \cc0.50 & 0.56 & 0.50 & 0.50 & 0.50 & 0.50  \\
Max    & \cc0.54 & 1.00 &  1.00 & 1.00 & 1.00 & 1.00  \\
\bottomrule
\end{tabular}
%\end{adjustbox}
}
}

\subtable[CelebA (10k)]{
%\begin{adjustbox}{max width=\columnwidth}
\resizebox{0.95\columnwidth}{!}{%
\begin{tabular}{c|cccccccc}
\toprule
$f$ & \cc $T$ & $0.975T$ & $0.875T$ & $0.75T$ & $0.625T$ & $0.5T$ \\
\midrule
Median & \cc0.94 & 0.89  & 0.83 & 0.67 & 0.54 & 0.50  \\
Sum    & \cc0.86 & 1.00 & 0.98 & 0.97 & 0.92 & 0.74  \\
Min    & \cc0.50 & 0.50 & 0.49 & 0.49 & 0.49 & 0.50 \\
Max    & \cc0.50 & 0.97 & 0.99 & 1.00 & 0.99 & 0.93 \\
\bottomrule
\end{tabular}
%\end{adjustbox}
}
}

\subtable[CelebA (15k)]{
%\begin{adjustbox}{max width=\columnwidth}
\resizebox{0.95\columnwidth}{!}{%
\begin{tabular}{c|ccccccccc}
\toprule
$f$ & \cc $T$ & $0.975T$ & $0.875T$ & $0.75T$ & $0.625T$ & $0.5T$ \\
\midrule
Median & \cc0.80 & 0.77 & 0.67 & 0.56 & 0.51 & 0.50 \\
Sum    & \cc0.74 & 0.96 & 0.98  & 0.95 & 0.81 & 0.62  \\
Min    & \cc0.50 & 0.50 & 0.50 & 0.50 & 0.50 & 0.50 \\
Max    & \cc0.50 & 0.81 & 0.99 & 0.99 & 0.96 & 0.80 \\
\bottomrule
\end{tabular}
%\end{adjustbox}
}
}
\subtable[CelebA (20k)]{
%\begin{adjustbox}{max width=\columnwidth}
\resizebox{0.95\columnwidth}{!}{%
\begin{tabular}{c|ccccccccc}
\toprule
$f$ & \cc $T$ & $0.975T$ & $0.875T$ & $0.75T$ & $0.625T$ & $0.5T$ \\
\midrule
Median & \cc 0.77 & 0.65 & 0.65 & 0.55 & 0.51 & 0.50  \\
Sum    & \cc 0.71 & 0.84 & 0.97 & 0.93 & 0.79 & 0.61 \\
Min    & \cc 0.50 & 0.49 & 0.50 & 0.50 & 0.50 & 0.50 \\
Max    & \cc 0.50  & 0.65 & 0.98 & 0.98 & 0.95 & 0.77 \\
\bottomrule
\end{tabular}
%\end{adjustbox}
}
}
\caption{The \textbf{white-box} attacks AUCROC with different statistic function $f$ and truncation steps $T_{trun}$ (shown in each column) on CelebA across various training set sizes (shown in the title of each sub-table). The first column $T$ corresponds to ``no truncation''.} 
\label{tab:wb_celeba_truncation_all}
\end{table}
We visualize the results of our white-box attack across various settings of the truncation steps on the CelebA dataset across various training configurations. We present the results in \figureautorefname~\ref{fig:white_vlb_truncation}. This is supplementary to the results in \tableautorefname~\ref{tab:whitebox_truncate_steps_exploration} in the main paper.

\begin{figure*}[!h]
\centering
\subfigure[White-box]{
\includegraphics[width=0.4\textwidth]{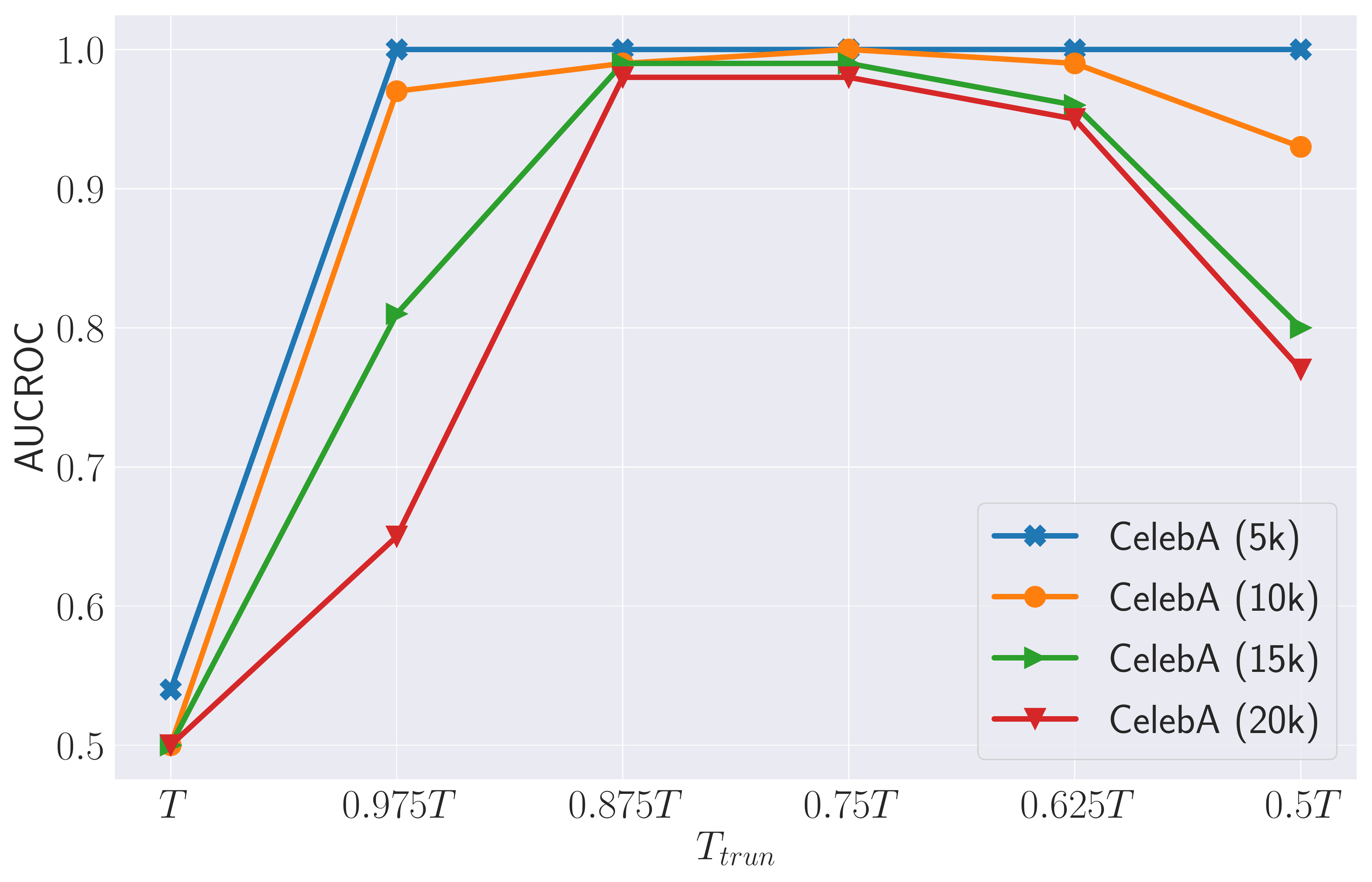}\label{fig:white_vlb_truncation}
}
\subfigure[Gray-box]{
\includegraphics[width=0.4\textwidth]{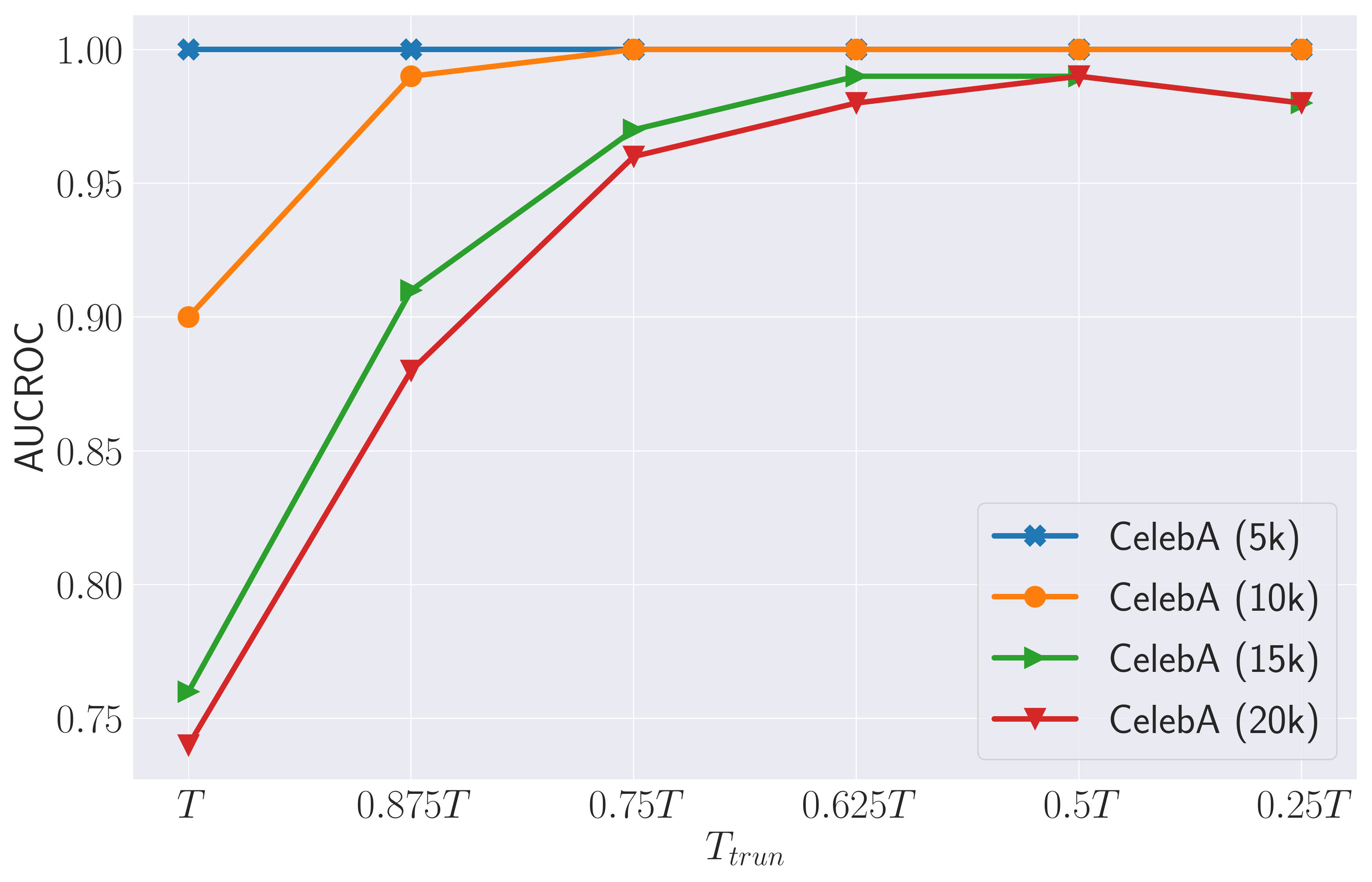}
\label{fig:greybox_mse_truncation}
}
\caption{The \ref{fig:white_vlb_truncation} \textbf{white-box} and \ref{fig:greybox_mse_truncation} \textbf{gray-box} attack AUCROC for different truncating steps $T_{trun}$ across different dataset sizes on CelebA. The statistic function is selected to be ``\textit{Max}'' and  ``\textit{Median}'' for the white-box and gray-box attack, respectively. }

\end{figure*}

% \begin{figure}[!h]
% \centering
% \includegraphics[width=0.9\columnwidth]{figures/greybox_mse_truncation.pdf}
%   \caption{The \textbf{gray-box} Attack AUCROC for different truncating steps $T_{trun}$ across different dataset sizes on CelebA. The statistic function is selected to be ``\textit{Median}''. }
% \label{fig:greybox_mse_truncation}
% \end{figure}

\subsection{Gray-box Setting}
\begin{table}[!h]
\aboverulesep=0ex
\belowrulesep=0ex
\centering
\definecolor{Gray}{gray}{0.9}
\newcommand{\cc}{\cellcolor{Gray}}
\centering
\subtable[CelebA (5k)]{
%\begin{adjustbox}{max width=\columnwidth}
\resizebox{0.95\columnwidth}{!}{%
\begin{tabular}{c|cccccc}
\toprule
$f$ & \cc $T$ & $0.875T$ & $0.725T$ & $0.625T$ & $0.5T$ & $0.25T$ \\
\midrule
Median & \cc1.00 & 1.00 & 1.00 & 1.00 & 1.00 & 1.00 \\
Sum   & \cc0.69 & 0.80 & 0.94 & 1.00 & 1.00 & 1.00 \\
Min    & \cc0.59 & 0.55 & 0.55 & 0.58 & 0.55 & 0.55 \\
Max    &\cc 0.50 & 0.50 & 0.54 & 0.66 & 1.00 & 1.00\\
\bottomrule
\end{tabular}
%\end{adjustbox}
}
}

\subtable[CelebA (10k)]{
%\begin{adjustbox}{max width=\columnwidth}
\resizebox{0.95\columnwidth}{!}{%
\begin{tabular}{c|cccccccc}
\toprule
$f$ & \cc $T$ & $0.875T$ & $0.725T$ & $0.625T$ & $0.5T$ & $0.25T$ \\
\midrule
Median & \cc0.90 & 0.99 & 1.00 & 1.00 & 1.00 & 1.00 \\
Sum    & \cc0.59 & 0.67 & 0.79 & 0.94 & 1.00 & 0.98 \\
Min    & \cc0.50 & 0.50 & 0.50 & 0.51 & 0.49 & 0.50 \\
Max    & \cc0.49 & 0.50 & 0.50 & 0.55 & 0.86 & 0.97 \\
\bottomrule
\end{tabular}
%\end{adjustbox}
}
}

\subtable[CelebA (15k)]{
%\begin{adjustbox}{max width=\columnwidth}
\resizebox{0.95\columnwidth}{!}{%
\begin{tabular}{c|ccccccccc}
\toprule
$f$ & \cc $T$ & $0.875T$ & $0.725T$ & $0.625T$ & $0.5T$ & $0.25T$ \\
\midrule
Median & \cc 0.76 & 0.91 & 0.97 & 0.99 & 0.99 & 0.99 \\
Sum    &\cc 0.55 & 0.60 & 0.70 & 0.85 & 0.96 & 0.99 \\
Min    & \cc0.50 & 0.50 & 0.50 & 0.50 & 0.50 & 0.50 \\
Max    & \cc0.50 & 0.50 & 0.50 & 0.52 & 0.76 & 1.00\\
\bottomrule
\end{tabular}
%\end{adjustbox}
}
}
\subtable[CelebA (20k)]{
%\begin{adjustbox}{max width=\columnwidth}
\resizebox{0.95\columnwidth}{!}{%
\begin{tabular}{c|ccccccccc}
\toprule
$f$ & \cc $T$ & $0.875T$ & $0.75T$ & $0.625T$ & $0.5T$ & $0.25T$ \\
\midrule
Median & \cc 0.74 & 0.88 & 0.96 & 0.98 & 0.99 & 0.98 \\
Sum   & \cc0.55 & 0.60 & 0.69 & 0.83 & 0.95 & 0.98 \\
Min    & \cc0.50 & 0.50 & 0.50 & 0.50 & 0.50 & 0.50 \\
Max    & \cc0.50 & 0.50 & 0.50 & 0.51 & 0.74 & 0.99\\
\bottomrule
\end{tabular}
%\end{adjustbox}
}
}
\caption{The \textbf{gray-box} attacks AUCROC with different statistic function $f$ and truncation steps $T_{trun}$ (shown in each column) on CelebA across various training set sizes (shown in the title of each sub-table). The first column $T$ corresponds to ``no truncation''.} 
\label{tab:gb_celeba_truncation_all}
\end{table}

%\vspace{6pt}
We provide additional quantitative results for our gray-box attack investigating the selection of truncation steps across various training set sizes on CelebA. The results are presented in \tableautorefname~\ref{tab:gb_celeba_truncation_all}, with the qualitative illustration being presented in \figureautorefname~\ref{fig:greybox_mse_truncation}.

\subsection{Black-box Setting}
\begin{figure*}[!h]
\centering
\subfigure[@0.1 FPR]{
\includegraphics[width=0.4\textwidth]{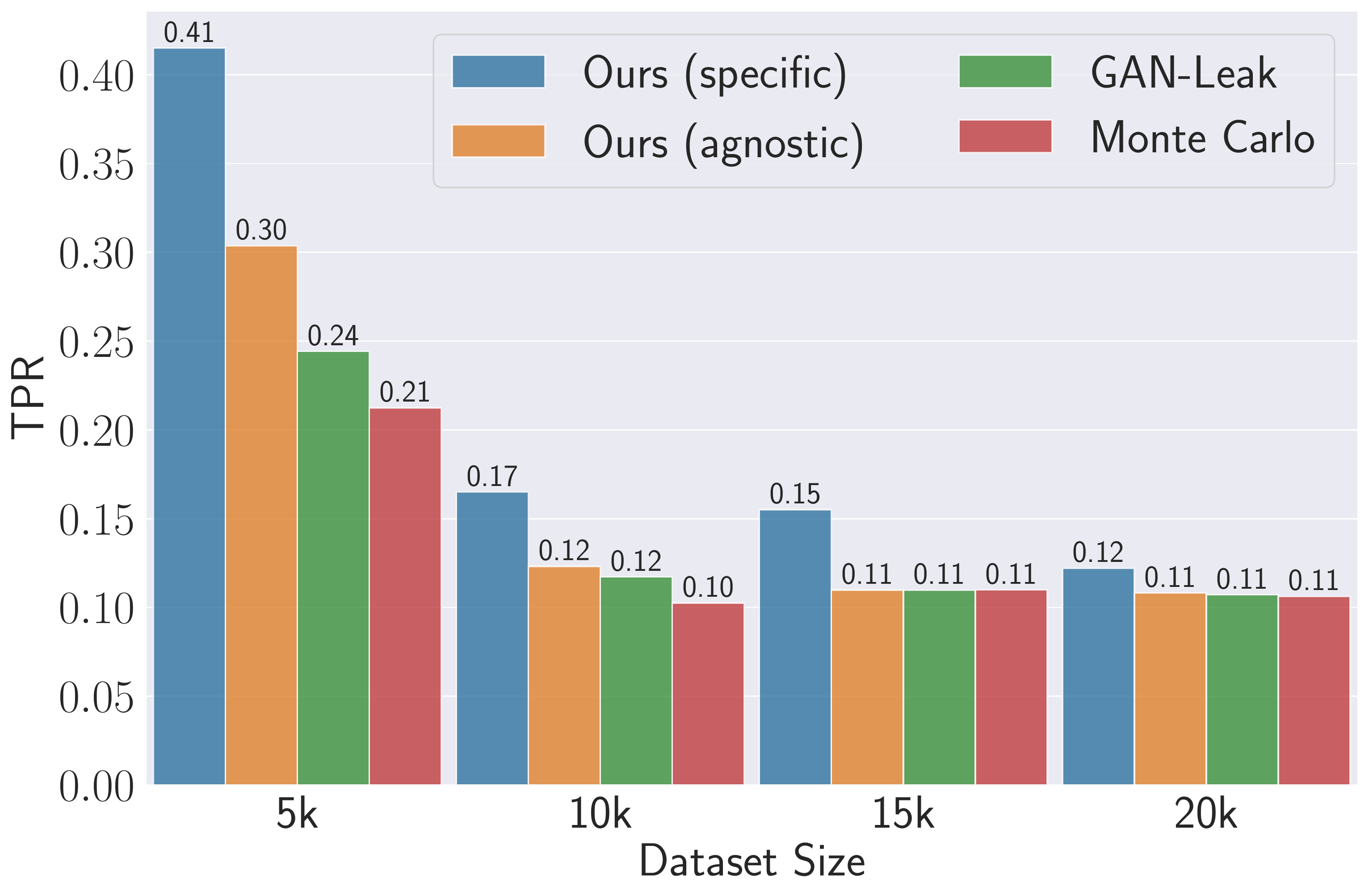}
}
\subfigure[@0.01 FPR]{
\includegraphics[width=0.4\textwidth]{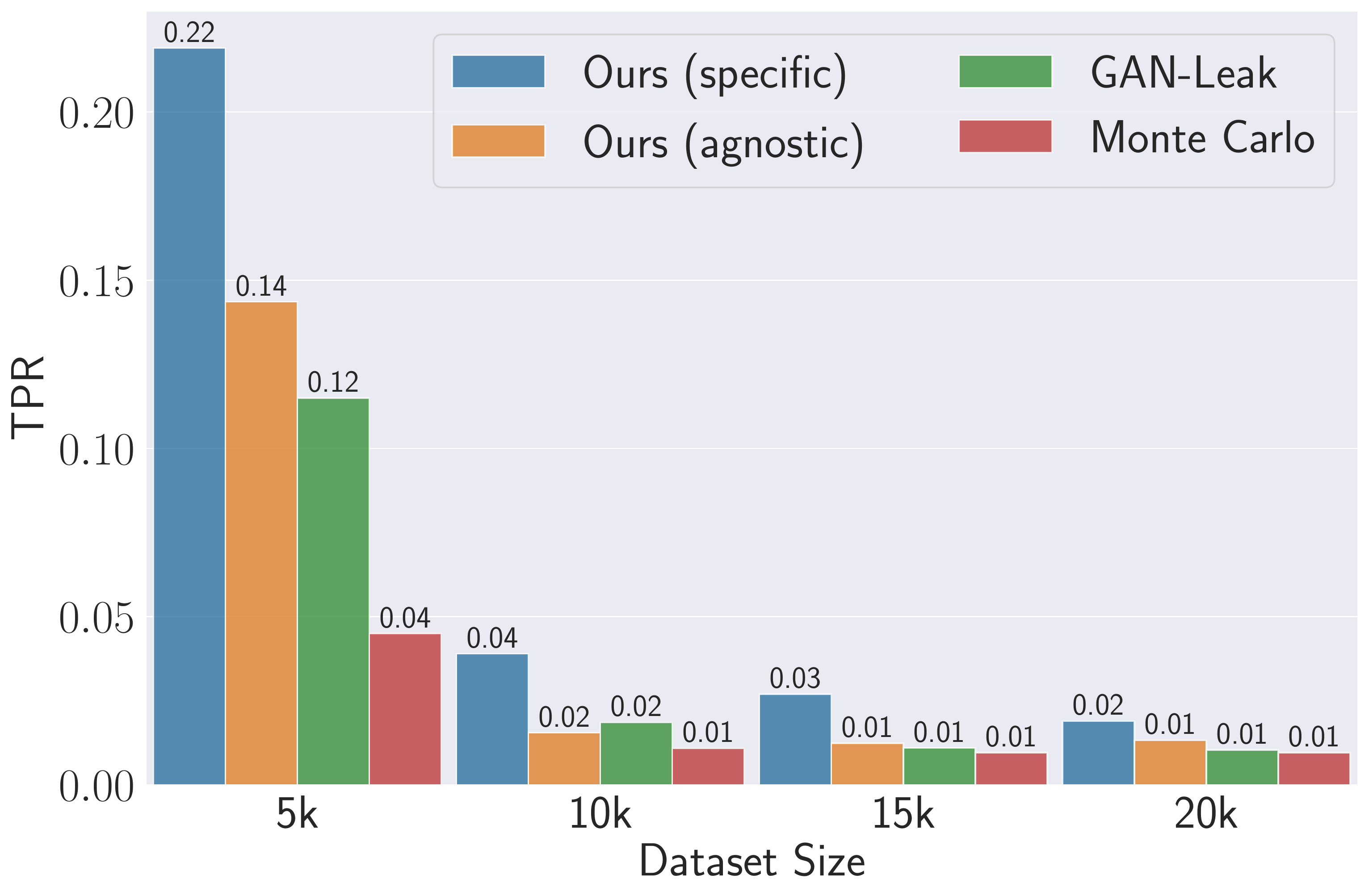}    
}
\caption{The black-box  TPR at low FPR on CIFAR-10.}
\label{fig:blackbox_TPR_cifar}
\end{figure*}

\begin{figure*}[!h]
\centering
\subfigure[@0.1 FPR]{
\includegraphics[width=0.4\textwidth]{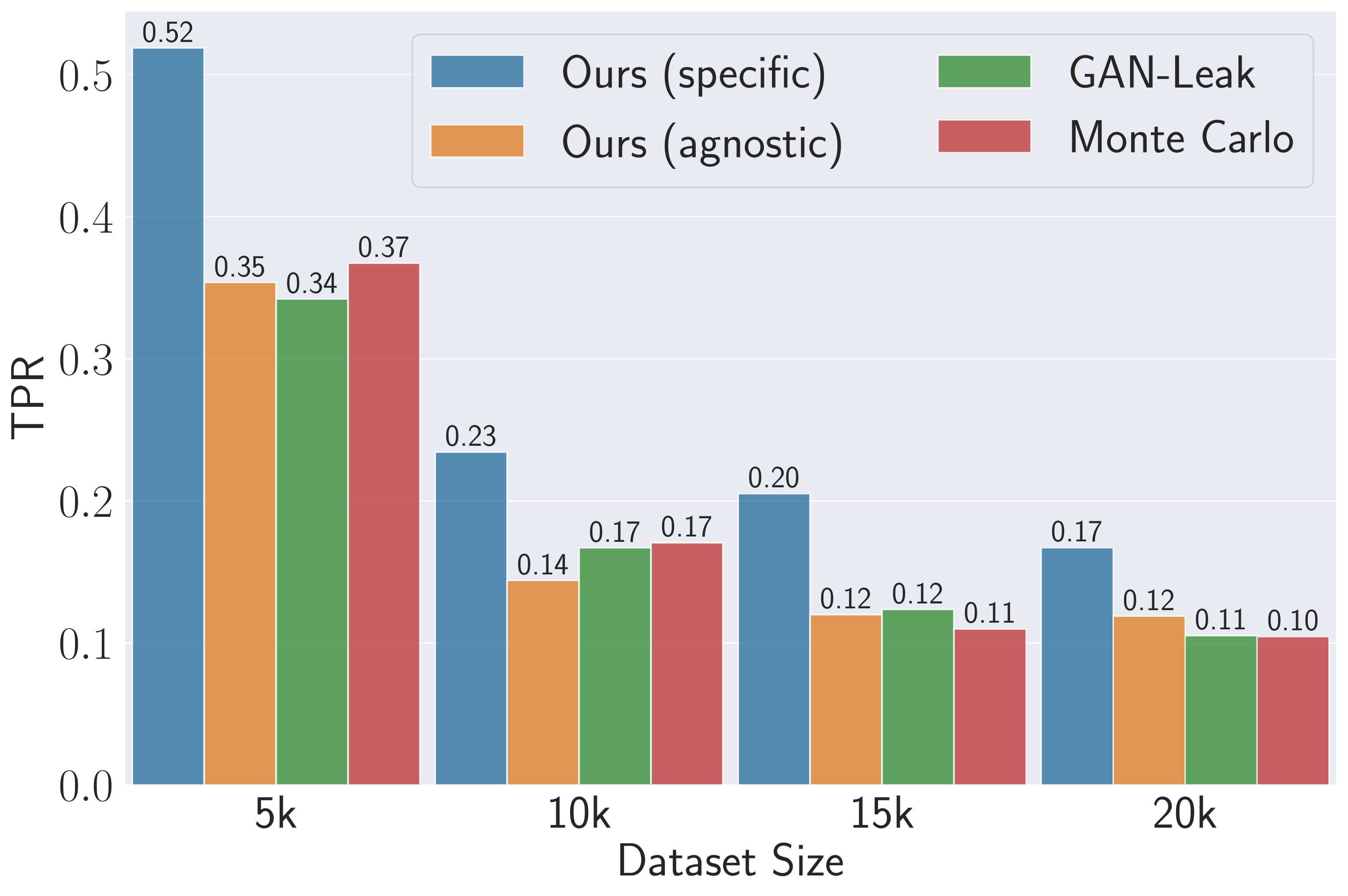}
}
\subfigure[@0.01 FPR]{
\includegraphics[width=0.4\textwidth]{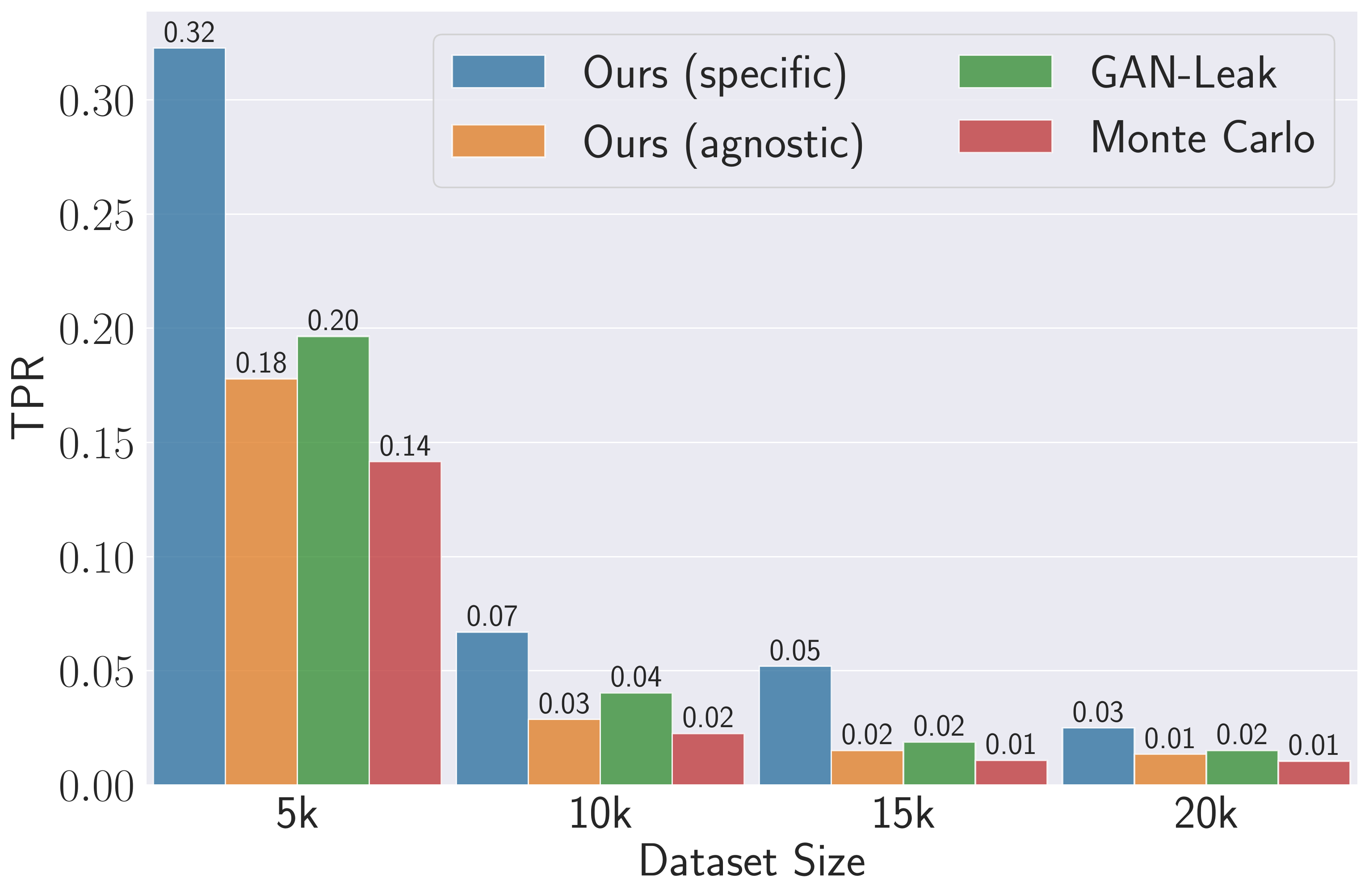}
}
\caption{The black-box  TPR at low FPR on CelebA.}
\label{fig:blackbox_TPR_celeba}
\end{figure*}

We present the TPR of different black-box attacks at certain levels of low FPR in Figure~\ref{fig:blackbox_TPR_cifar} and Figure~\ref{fig:blackbox_TPR_celeba}, supplementing the results in Figure~\ref{fig:blackbox_comparison} in the main paper. As can be seen from the plots, the comparison results over TPR are largely consistent with those obtained using AUCROC, though our model-specific attack shows a greater advantage over others when compared at TPR@(0.1 or 0.01)FPR. Consistently across all configurations, the TPR achieved by our attack is higher than the FPR, indicating a successful attack due to its ability to more accurately identify members than incorrectly predict non-members as members. Importantly, an MIA can be regarded as successful if it can reliably identify even a few members.

\begin{figure*}[!h]
\centering
\subfigure[@0.1 FPR]{
\includegraphics[width=0.4\textwidth]{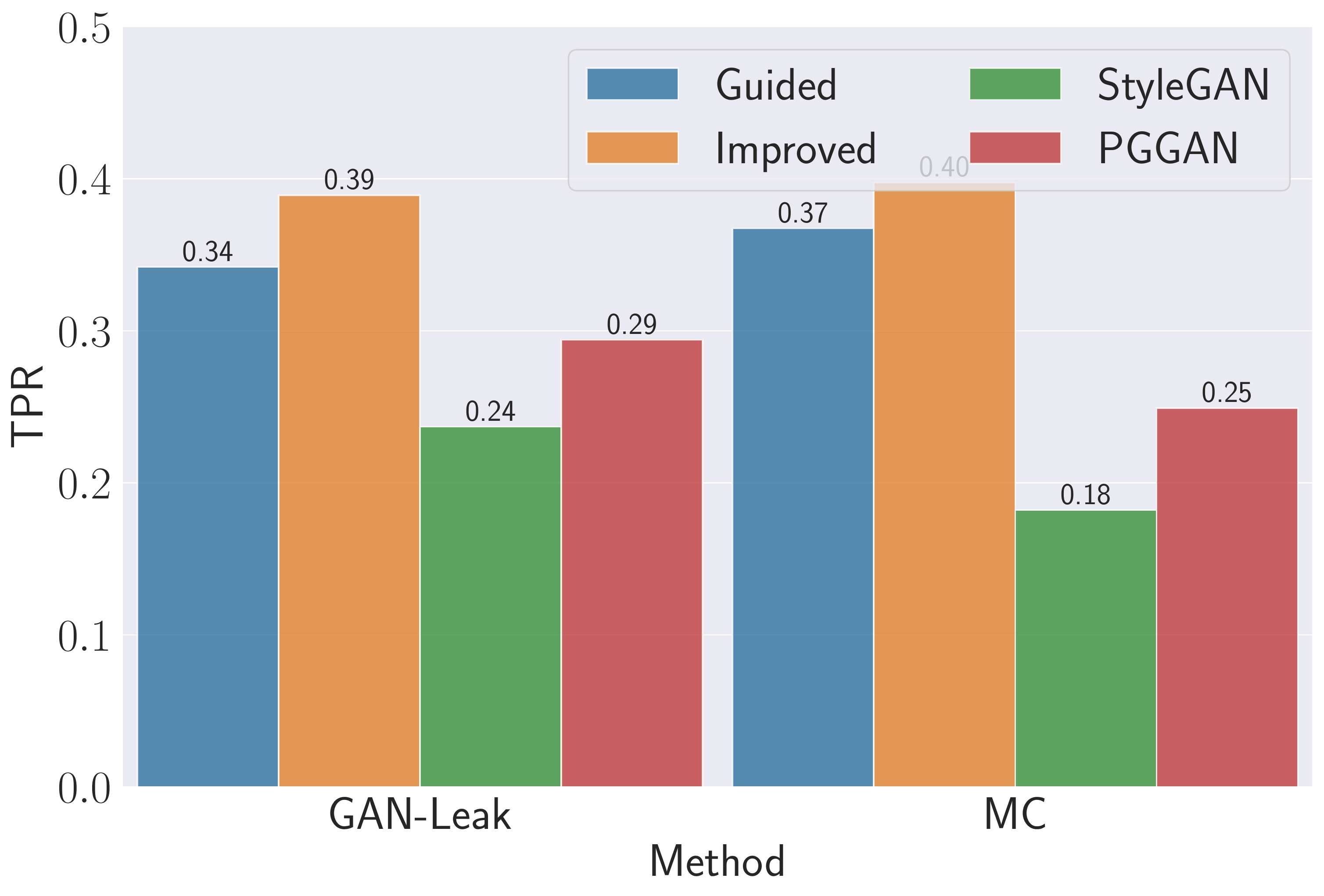}
}
\subfigure[@0.01 FPR]{
\includegraphics[width=0.4\textwidth]{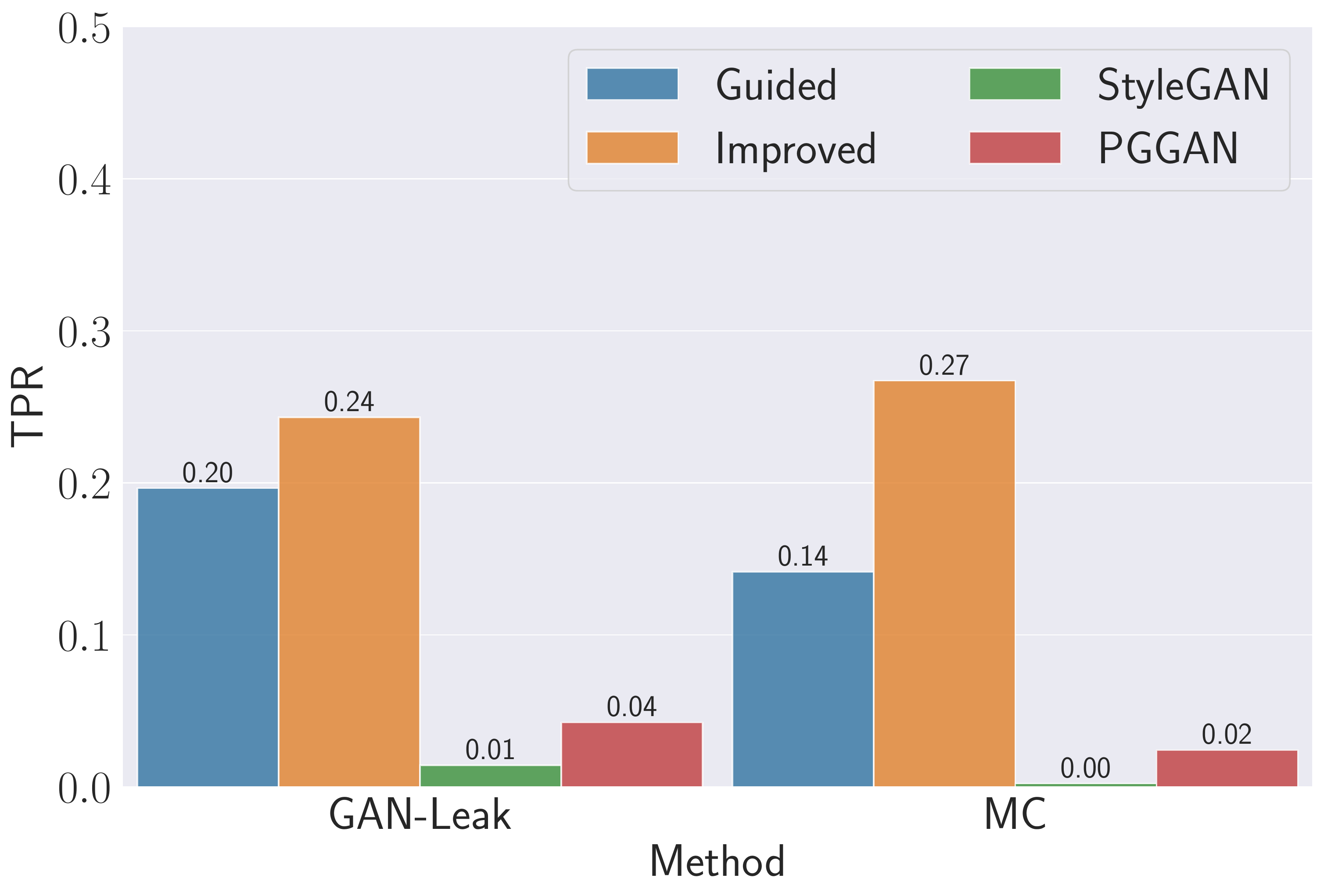}        
}
\caption{The model-agnositc black-box TPR at low FPR for various generative models on CelebA(5k).}
\label{fig:blackbox_TPR_gan_diffusion}
\end{figure*}

The TPR of different model-agnostic black-box attacks at certain levels of low FPR on various types of generative models is presented in Figure~\ref{fig:blackbox_TPR_gan_diffusion}, which supplements the results in Table~\ref{tab:comparison_gan_diffusion} in the main paper. The comparison results are consistent regardless of the adopted metrics (AUCROC or TPR) for evaluating attack performance, all showing that diffusion models generally have a higher vulnerability to MIA than GANs do, even when all of them are trained in the same controlled environment and exhibit similar generation quality (see the quantitative results in \tableautorefname~\ref{tab:comparison_gan_diffusion}).


\end{document}